\title{Dual Illumination Estimation for Robust Exposure Correction}

\author[Q. Zhang et al.]
{\parbox{\textwidth}{\centering Qing Zhang$^{1}$, Yongwei Nie${^2}$, and Wei-Shi Zheng$^{*,1}$
	}
	\\
	{\parbox{\textwidth}{\centering $^1$School of Data and Computer Science, Sun Yat-sen University, Guangzhou, China\\
	$^2$School of Computer Science and Engineering, South China University of Technology, Guangzhou, China
	}
}
}

\begin{document}
\teaser{
	 \centering
 \includegraphics[height=1.09in]{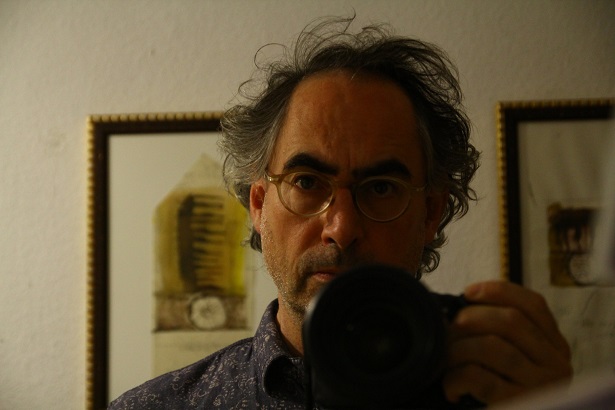}
  \includegraphics[height=1.09in]{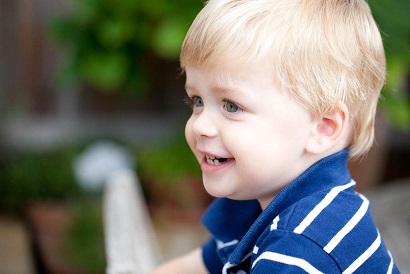} 
  \includegraphics[height=1.09in]{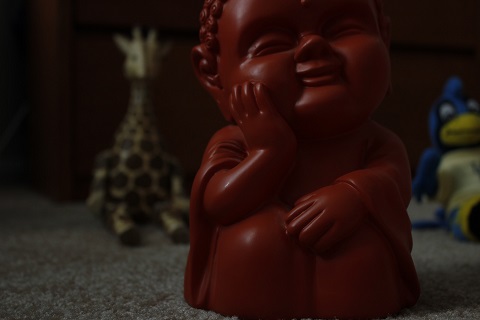}
  \includegraphics[height=1.09in]{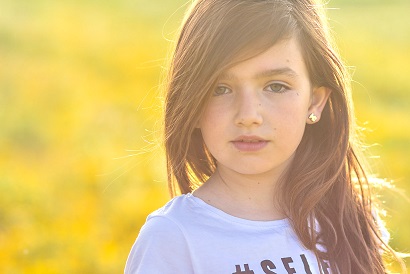}  \\ \vspace{2pt}
   \includegraphics[height=1.09in]{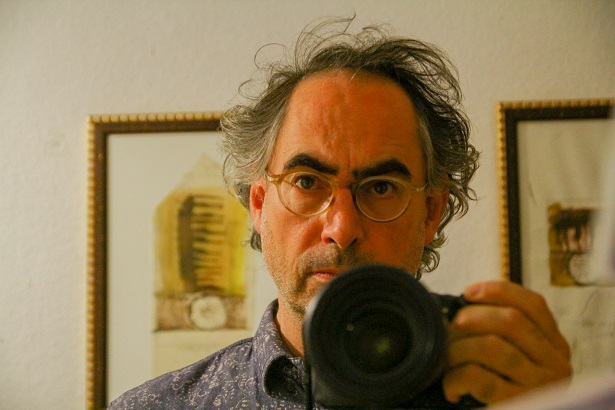}
 \includegraphics[height=1.09in]{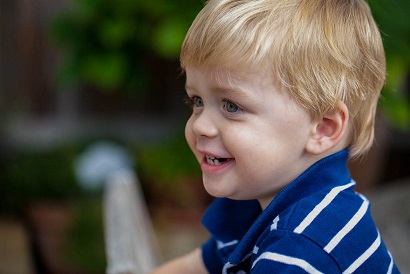}
 \includegraphics[height=1.09in]{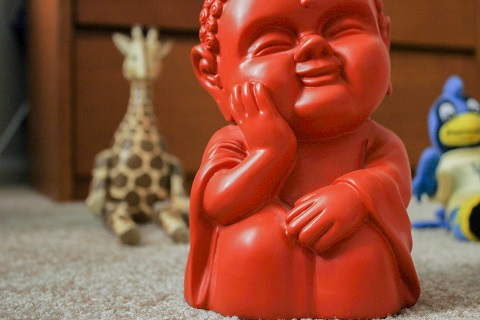}
 \includegraphics[height=1.09in]{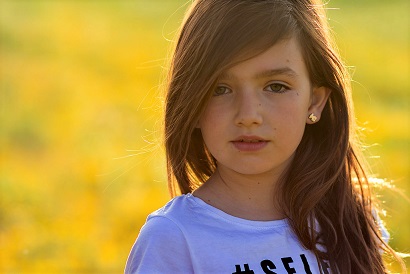} \\ 

  \caption{Some example results produced by our method. The input images are on the top and involve diverse exposure conditions. Our results are on the bottom. Note the improved brightness, distinct contrast, clear details, and vivid color in our results.}
  \vspace{1mm}
  	
\label{fig:teaser}
}

\maketitle
\begin{abstract}
Exposure correction is one of the fundamental tasks in image processing and computational photography. While various methods have been proposed, they either fail to produce visually pleasing results, or only work well for limited types of image (\eg, underexposed images). In this paper, we present a novel automatic exposure correction method, which is able to robustly produce high-quality results for images of various exposure conditions (\eg, underexposed, overexposed, and partially under- and over-exposed). At the core of our approach is the proposed dual illumination estimation, where we separately cast the under- and over-exposure correction as trivial illumination estimation of the input image and the inverted input image. By performing dual illumination estimation, we obtain two intermediate exposure correction results for the input image, with one fixes the underexposed regions and the other one restores the overexposed regions. A multi-exposure image fusion technique is then employed to adaptively blend the visually best exposed parts in the two intermediate exposure correction images and the input image into a globally well-exposed image. Experiments on a number of challenging images demonstrate the effectiveness of the proposed approach and its superiority over the state-of-the-art methods and popular automatic exposure correction tools.
%
%
%
\end{abstract}  
\section{Introduction}
\footnotetext{*Corresponding author.}
With the prevalence of camera-embedded mobile devices and inexpensive digital cameras, people are increasingly interested in taking photos, so that photo sharing on social networks has become a trendy lifestyle. However, despite modern cameras are equipped with many sophisticated techniques and are generally easy to control and use, capturing well-exposed photos under complex lighting conditions (\eg, low light and back light) remains a challenge for non-professional photographers. 
Hence, poorly exposed photos are inevitably created; see Figure~\ref{fig:teaser} for examples. Due to the unclear details, weak contrast and dull color, such photos usually look unpleasing and fail to capture user-desired effects, which increases the need for effective exposure correction techniques. 


\begin{figure*}
	\centering
	\begin{subfigure}[c]{0.19\textwidth}
		\includegraphics[width=1.32in]{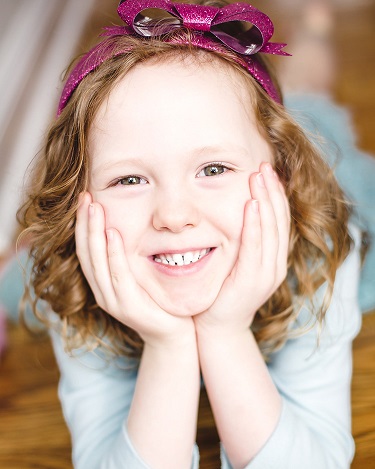}
		\caption{Input ($750 \times 938$)}
	\end{subfigure}
	\begin{subfigure}[c]{0.19\textwidth}
		\includegraphics[width=1.32in]{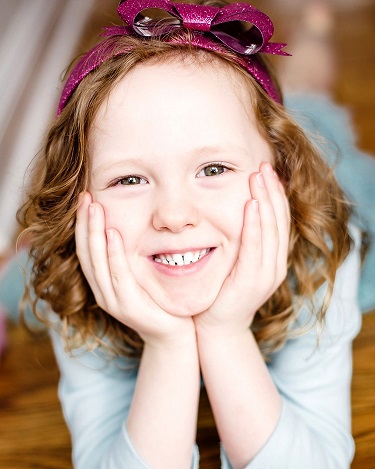}
		\caption{Auto-Level ($<0.1$s)}
	\end{subfigure}
	\begin{subfigure}[c]{0.19\textwidth}
		\includegraphics[width=1.32in]{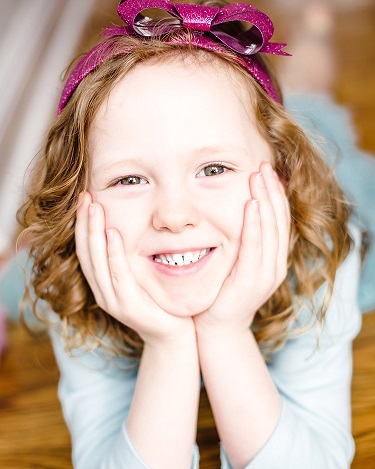}
		\caption{Auto-Tone ($<0.1$s)}
	\end{subfigure}
	\begin{subfigure}[c]{0.19\textwidth}
		\includegraphics[width=1.32in]{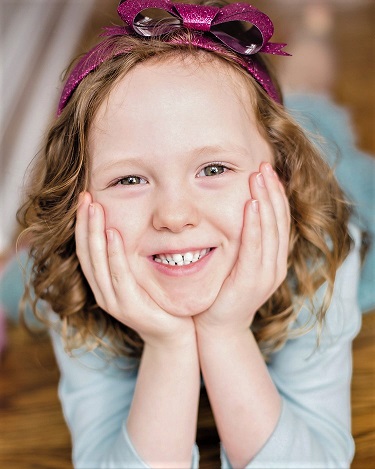}
		\caption{Expert-retouched (8s)}
	\end{subfigure}
	\begin{subfigure}[c]{0.19\textwidth}
		\includegraphics[width=1.32in]{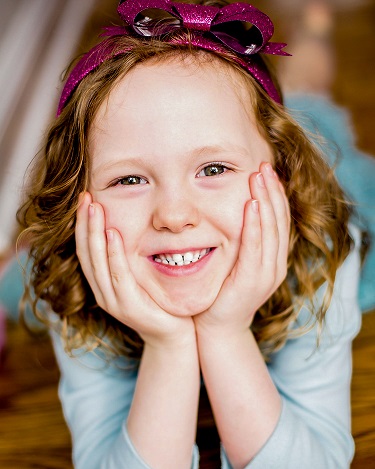}
		\caption{Our result (0.8s)}
	\end{subfigure}
	\vspace{-2mm}
	\caption{An overexposed image processed by various exposure correction tools. (b) and (c) are results generated by Auto-Level in Photoshop and Auto-Tone in Lightroom, while result (d) is produced by a Photoshop expert through interactive adjustment. The time cost for generating each result is also shown for evaluating the ease of use and algorithm efficiency.}
	\label{fig:compare_tools}
	\vspace{-2mm}
\end{figure*}

Because of the inherent nonlinearity and subjectivity, exposure correction is a challenging task. Indeed, existing image editing softwares (\eg, Photoshop, GIMP and Lightroom) offer various tools for users to interactively adjust the tone and exposure of photos, while they remain difficult for non-experts since these tools basically require a tedious process to balance multiple controls (\eg, brightness, contrast, color, etc.). Although the ``Auto-Tone'' feature in Lightroom and the ``Auto-Level'' feature in Photoshop allow automatic exposure correction by just a single click, they may not always apply the right adjustments to the input image, making them fail to produce satisfactory results. Figure~\ref{fig:compare_tools} shows an example image processed by these tools.

Researchers have also developed various exposure correction methods. However, they are mostly designed for solely correcting under-~\cite{wang2013naturalness,guo2017lime,zhang2018high} or over-exposure~\cite{guo2010correcting,lee2014correction,abebe2018towards}, thereby having limited applicability. There also exists some methods that are applicable to images of arbitrary exposure conditions. Early methods such as histogram equalization and its variants~\cite{zuiderveld1994contrast,kim1997contrast,stark2000adaptive} work by stretching the dynamic range of the intensity histogram, but tend to generate unrealistic results. Some subsequent methods rely on S-shaped tone mapping curves~\cite{reinhard2002photographic, yuan2012automatic} or wavelet~\cite{hou2013recovering} to work, while more recent methods~\cite{gharbi2017deep,chen2018deep,hu2018exposure,wang2019underexposed} train tone adjustment models on datasets to allow exposure correction. However, they do not work well on overexposed images and may induce unnatural results; see Figure~\ref{fig:user_study_2}. 


This paper presents a novel exposure correction approach, which is built upon the observation that under- and over-exposure correction can be jointly formulated as a trivial illumination estimation problem of the input image and the inverted input image. Although previous methods have demonstrated the effectiveness of illumination estimation in correcting underexposed photos, they barely explore its potential in handling overexposure. Unlike them, we found that overexposure correction can also be formulated as an illumination estimation problem by inverting the input image, since the originally overexposed regions would appear as underexposed, allowing us to fix overexposed regions in the input image by correcting underexposed regions in the inverted input image. Hence, we introduce dual illumination estimation, where we separately predict \emph{forward illumination} for the input image and \emph{reverse illumination} for the inverted input image. Two intermediate exposure correction images of the input image are then recovered from the estimated forward and reverse illuminations, with one that fixes the underexposed regions and the other one restores the overexposed regions. Next, we apply an effective multi-exposure image fusion to the intermediate exposure correction images and the input image to seamlessly blend the locally best exposed parts in each of the three images into a globally well-exposed image. 


The contribution of this paper is a simple yet effective exposure correction method built upon a novel dual illumination estimation. To show the effectiveness of our method, we evaluate it on a number of challenging images and compare it against both state-of-the-art methods and the popular exposure correction tools via user study. Experiments show that results generated by our method are more preferred by human subjects, and our method is effective to deal with previously challenging images (\eg, images with both under- and over-exposed regions). Moreover, our method is fully automatic and can run at near-interactive rate. 

\section{Related Work}
Exposure correction is an important research problem with an immense literature. Perhaps, the most fundamental method is histogram equalization (HE), which globally enhances image contrast by stretching the intensity histogram. Despite the simplicity and effectiveness in contrast enhancement, it tends to generate unrealistic results because of ignoring relationship between pixels. 


Mertens~\etal~\cite{mertens2009exposure} proposed to blend well-exposed regions from an image sequence with bracketed exposure into a single high-quality image. Despite the success of this technique, it cannot be directly applied to a single image because of requiring multi-exposure image sequence as input. Later, Zhang~\etal~\cite{zhang2016underexposed} adapted this technique to underexposed video enhancement by first constructing multi-exposure image sequence for each video frame using sampled tone mapping curves, and then obtained the enhanced video by progressively fusing the image sequences in a spatio-temporal aware fashion. In contrast, our method can not only work for a single image, but also is applicable to images of various exposure conditions, not just underexposed images.

\begin{figure*}
	\centering
	\includegraphics[width=6.7in]{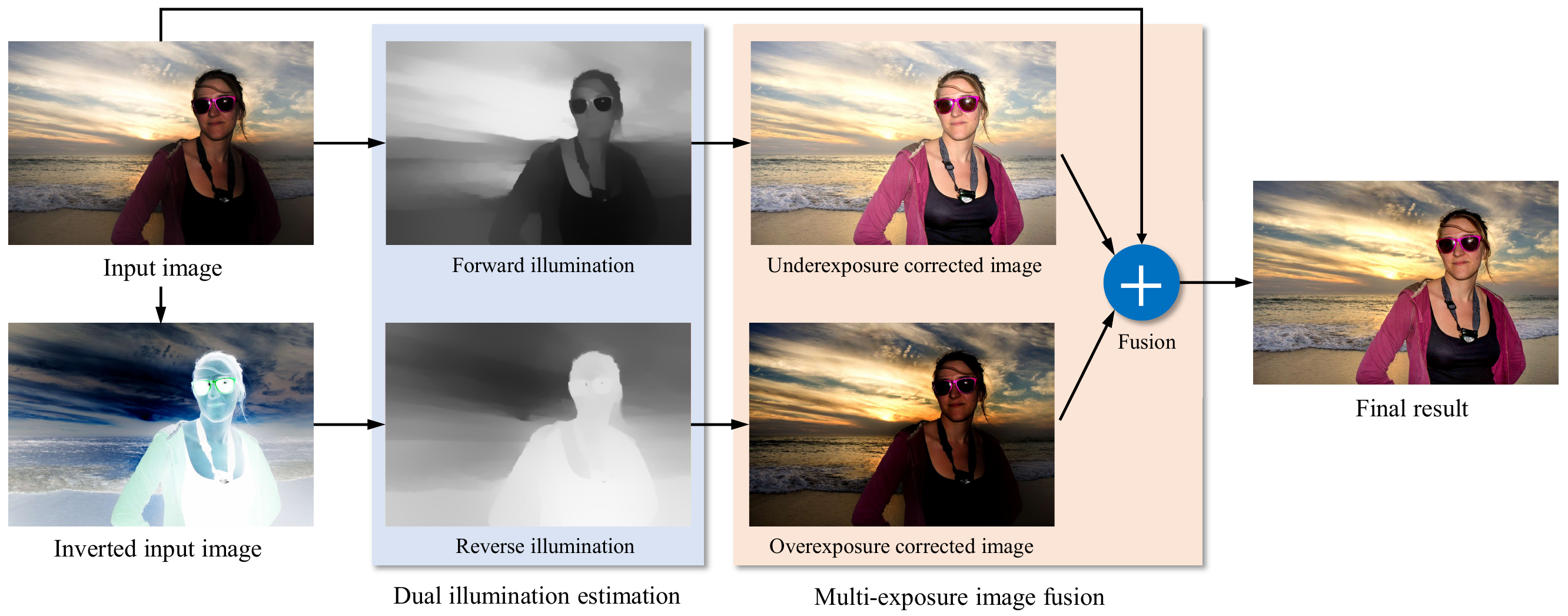}
	\vspace{-2mm}
	\caption{Overview of the proposed exposure correction algorithm. Given an input image, the dual illumination estimation is first performed to obtain the forward and reverse illuminations, from which we then recover the intermediate under- and over-exposure corrected images of the input. Next, an effective multi-exposure image fusion is applied to seamlessly blend visually best exposed parts in the two intermediate exposure correction images as well as the input image into the final globally well-exposed image. }
	\label{fig:overview}
	\vspace{-2mm}
\end{figure*}

Bennett and McMillan~\cite{bennett2005video} decomposed an input image into base and detail layers, and applied different sigmoid mappings for the two layers to restore the underexposed regions while preserving the image details. However, it often produces over saturated results. Yuan and Sun~\cite{yuan2012automatic} described an automatic exposure correction method for consumer photographs. The key idea behind their method is to infer optimal exposure for each subregion and map subregions to their desired exposure levels using detail-preserving tone mapping curve. While this method demonstrates promising results, it may also fail because it relies on reliable region segmentation, which is a challenging task.

Reverse tone mapping can also be used to correct exposure of an image by inferring a high dynamic range (HDR) image from a single low dynamic range (LDR) input  \cite{masia2016content,endo2017deep,eilertsen2017hdr}. Our work differs them in two aspects. First, we do not change the bit depth of the input image. Second, our results can be displayed in any devices, without an additional tone mapping operation.

Interactive exposure correction methods were also developed. Lischinski~\etal~\cite{lischinski2006interactive} presented an interactive tool for tone adjustment. Given an input image, they allow users to quickly select the regions of interest by drawing a few brush strokes and then locally adjust the brightness, contrast, and other appearance factors in the selected regions through a group of sliders. Dodgson~\etal~\cite{dodgson2009contrast} introduced contrast brushes, an interactive method for contrast enhancement. Such methods benefit from user interactions, while our approach is fully automatic.

Content-aware methods utilize high-level image contents to acquire better exposure correction effects for areas of interest, \eg, human faces, skin and sky, etc. Joshi~\etal~\cite{joshi2010personal} improved the quality of faces in personal photo collection by using high-quality photos of the same person as examples. Dale~\etal~\cite{dale2009image} presented an example-based image restoration method that leverages a large database of Internet images. Kaufman~\etal~\cite{kaufman2012content} described a photo enhancement framework that takes both local and global image semantics into account. A common limitation of these methods is that they are usually very sensitive to the reliability of the extracted image semantics.


\begin{figure}
	\centering
	\begin{subfigure}[c]{0.22\textwidth}
		\centering
		\includegraphics[width=1.52in]{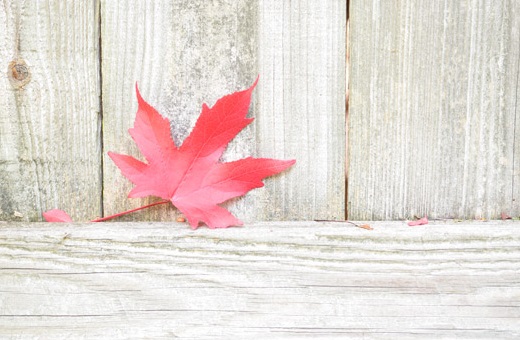}
		\caption{Input}
	\end{subfigure}
	\begin{subfigure}[c]{0.22\textwidth}
		\centering
		\includegraphics[width=1.52in]{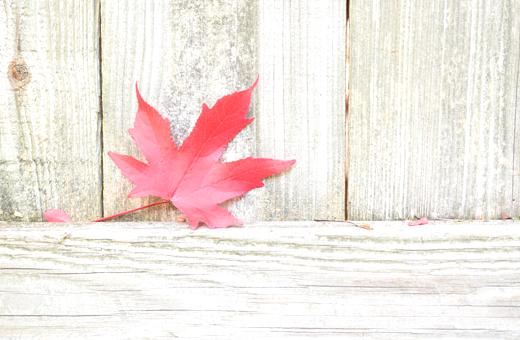}
		\caption{Guo~\etal~\cite{guo2017lime}}
	\end{subfigure}  \\ \vspace{1mm}
	\begin{subfigure}[c]{0.22\textwidth}
		\centering
		\includegraphics[width=1.52in]{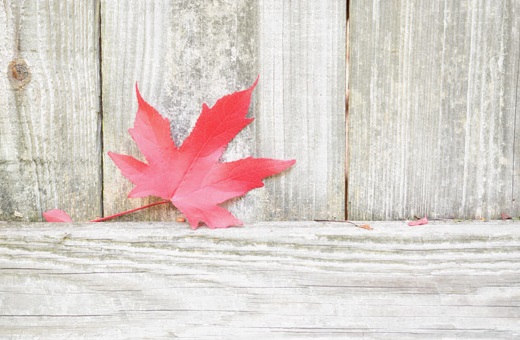}
		\caption{Zhang~\etal~\cite{zhang2018high}}
	\end{subfigure}
	\begin{subfigure}[c]{0.22\textwidth}
		\centering
		\includegraphics[width=1.52in]{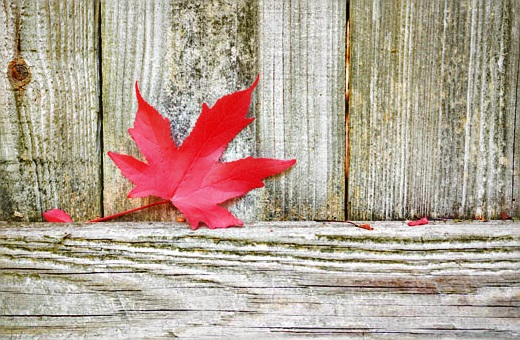}
		\caption{Our result}
	\end{subfigure} 
	\vspace{-2mm}
	\caption{Limitation of existing Retinex-based image enhancement methods (b) and (c) in correcting an overexposed image (a).}
	\label{fig:retinex_issue}
	\vspace{-2mm}
\end{figure}

\begin{figure*}
	\centering
	\begin{subfigure}[c]{0.19\textwidth}
		\centering
		\includegraphics[width=1.32in]{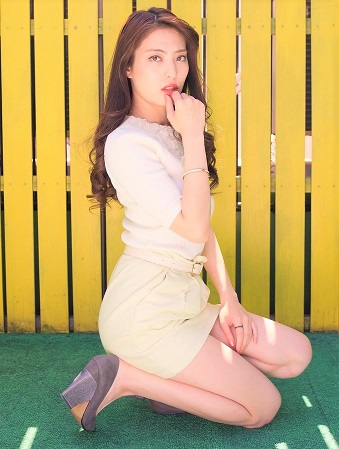}
		\caption{}
	\end{subfigure}
	\begin{subfigure}[c]{0.19\textwidth}
		\centering
		\includegraphics[width=1.32in]{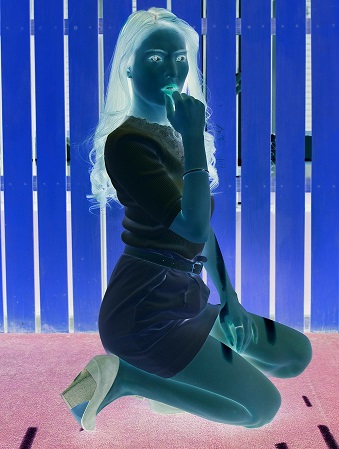}
		\caption{}
	\end{subfigure}
	\begin{subfigure}[c]{0.19\textwidth}
		\centering
		\includegraphics[width=1.32in]{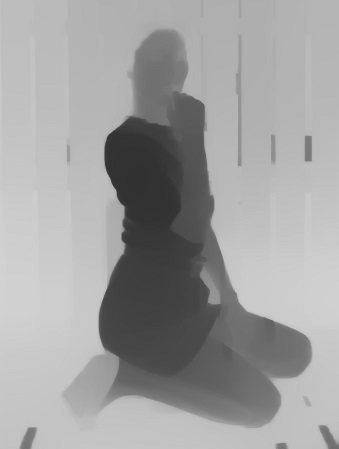}
		\caption{}
	\end{subfigure}
	\begin{subfigure}[c]{0.19\textwidth}
		\centering
		\includegraphics[width=1.32in]{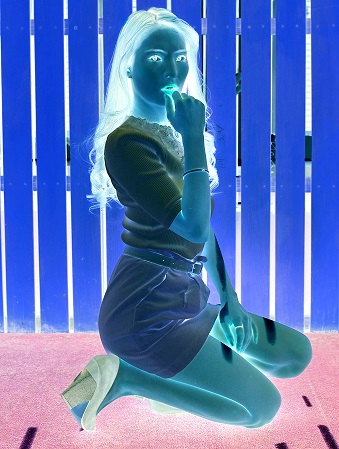}
		\caption{}
	\end{subfigure}
	\begin{subfigure}[c]{0.19\textwidth}
		\centering
		\includegraphics[width=1.32in]{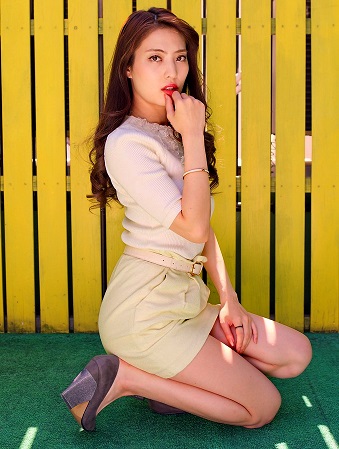}
		\caption{}
	\end{subfigure}	
	\vspace{-2mm}
	\caption{Validation of our observation. (a) Input overexposed image $I$. (b) Inverted input image $I_{inv}$. (c) and (d) are illumination $L_{inv}$ and underexposure corrected image $I'_{inv}$ of the inverted image $I_{inv}$. (e) Overexposure corrected image $I' = 1 - I'_{inv}$ of the input image (a).} 
	\label{fig:validate_observation}
	\vspace{-2mm}
\end{figure*}

Since the pioneering work of Bychkovsky~\etal~\cite{bychkovsky2011learning} who provided a dataset consisted of image pairs for tone adjustment, there is an increasing number of learning-based exposure correction methods. Yan~\etal~\cite{yan2016automatic} described a semantic-aware photo enhancement network. Cai~\etal~\cite{cai2018learning} learned a contrast enhancer from multi-exposure images by constructing a training dataset of low-contrast and high-contrast image pairs for end-to-end CNN (convolutional neural networks) learning, while Chen~\etal~\cite{chen2018deep} designed an unpaired learning model based on generative adversarial networks (GANs). Reinforcement learning was also employed to train photo adjustment models. For instance, Hu~\etal~\cite{hu2018exposure} achieved a white-box photo post-processing framework by modeling retouching operations as differential filters. Unlike~\cite{hu2018exposure}, Park~\etal~\cite{park2018distort} casted enhancement as a Markov Decision Process of several fundamental global color adjustment actions, and trained an agent on unpaired data to reveal the optimal sequence of actions. The limitation of learning-based methods is that they do not work well on images that are significantly different with the training images.





\section{Our Approach}

Figure~\ref{fig:overview} presents the system overview of our exposure correction algorithm. Given an input image, we first perform dual illumination estimation to obtain the forward and reverse illuminations, from which we recover the intermediate under- and over-exposure corrected images. Then, the two intermediate exposure correction images together with the input image are fused into the desired image that seamlessly blends the best exposed parts in each of the three images. In the following, we elaborate the proposed approach. Specifically, we first describe the dual illumination estimation (Section 3.1). Next, we present the multi-exposure image fusion (Section 3.2). Finally, we illustrate the implementation details and parameter setting (Section 3.3).

\subsection{Dual illumination estimation}
\textbf{Background.} Fundamental to our dual illumination estimation is the assumption in Retinex-based image enhancement~\cite{wang2013naturalness,guo2017lime,zhang2018high}, which assumes that an image $I$ (normalized to [0,1]) can be characterized as a pixel-wise product of the desired enhanced image $I'$ and a single-channel illumination map $L$:
\begin{equation} \label{equ:retinex}
I = I' \times L,
\end{equation}
where $\times$ denotes pixel-wise multiplication. With this assumption, image enhancement can be reduced to an illumination estimation problem, since we can recover the desired image $I'$ as long as the illumination map $L$ is known. However, Retinex-based methods do not work well on overexposed images. The reason is that attenuating exposure of an image require the illumination map $L$ in Eq.~\ref{equ:retinex} to exceed the normal gamut (\ie, $L > 1$), since the resulting image $I'$ is recovered by $I \times L^{-1}$. Figure~\ref{fig:retinex_issue} shows an example, where the Retinex-based enhancement methods further increase the exposure of the overexposed input image, generating visually unpleasing images in Figure~\ref{fig:retinex_issue}(b) and (c).


\noindent \textbf{Key observation.} Unlike previous Retinex-based enhancement methods, we observed that overexposure correction can also be formulated as an illumination estimation problem by inverting the input image, since the originally overexposed regions would appear as underexposed in the inverted image, allowing us to fix overexposed regions in the input image by correcting corresponding underexposed regions in the inverted input image. Specifically, to correct overexposed regions in an input image $I$, we first obtain its inverted image $I_{inv} = 1 - I$ and estimate the corresponding illumination map $L_{inv}$. We then compute underexposure corrected image $I'_{inv}$ by $I'_{inv} = I_{inv} \times L_{inv}^{-1}$ and recover the desired overexposure corrected image $I' = 1 - I'_{inv}$. Note, the inverted input image is usually an unrealistic image, but the recovered overexposure corrected image is realistic. Figure~\ref{fig:validate_observation} validates our observation, where we successfully correct an overexposed image by performing illumination estimation on the inverted input image. 

It is worth noting that inverted image has been utilized in previous enhancement methods \cite{dong2011fast,li2015low}. The differences between our use of inverted image and these methods are twofold. First, they focus on enhance low-light image/video, while we aim to correct overexposed photos. Second, their observation is that inverted low-light images look like hazy image, and thus dehazing algorithm is employed to produce the final results. In contrast, we observed that overexposed images are underexposed when inverted and can be indirectly corrected by illumination estimation.

Based on this observation, we design the dual illumination estimation, where the first pass estimates forward illumination for the input image and aims to correct underexposed regions, while the other pass is performed on the inverted input image for obtaining reverse illumination and correcting overexposed regions. The reason behind this design is that the input image may be partially under- and over-exposed, thus requiring two-pass illumination estimation to correct regions of different exposure conditions. Note the forward and reverse illuminations are separately estimated in the same illumination estimation framework. Below we describe the illumination estimation framework.

\begin{figure*}
	\centering
	\begin{subfigure}[c]{0.19\textwidth}
		\includegraphics[width=1.32in]{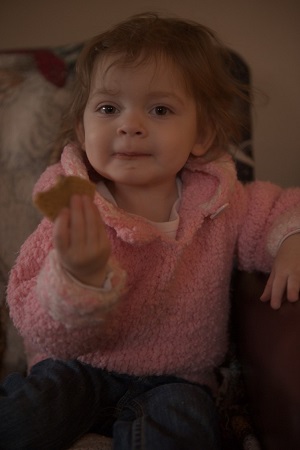}
		\vspace{-1.5em}
		\caption{}
	\end{subfigure}
	\begin{subfigure}[c]{0.19\textwidth}
		\includegraphics[width=1.32in]{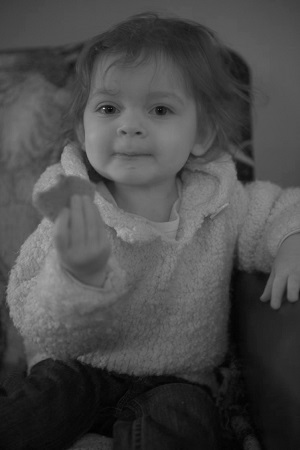}
		\vspace{-1.5em}
		\caption{}
	\end{subfigure}
	\begin{subfigure}[c]{0.19\textwidth}
		\includegraphics[width=1.32in]{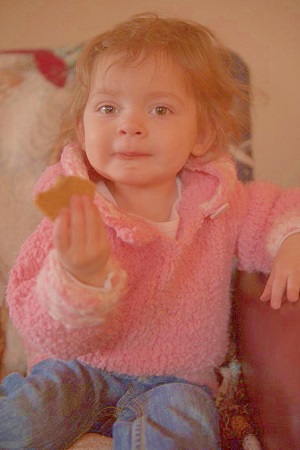}
		\vspace{-1.5em}
		\caption{}
	\end{subfigure}
	\begin{subfigure}[c]{0.19\textwidth}
		\includegraphics[width=1.32in]{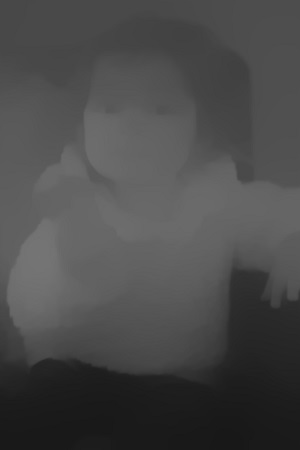}
		\vspace{-1.5em}
		\caption{}
	\end{subfigure}
	\begin{subfigure}[c]{0.19\textwidth}
		\includegraphics[width=1.32in]{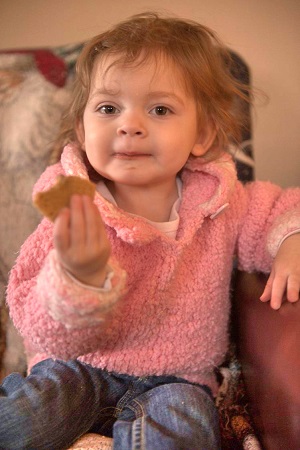}
		\vspace{-1.5em}
		\caption{}
	\end{subfigure}
	\vspace{-2mm}
	\caption{Illumination estimation. (a) Input image. (b) Initial illumination. (c) Result recovered from the initial illumination (b). (d) Our refined illumination. (e) Result recovered from our refined illumination (d). Note the forward illumination is estimated here, since the input image is obviously underexposed. Source image from Bychkovsky~\etal~\cite{bychkovsky2011learning}.}
	\label{fig:lum_estimation}
	\vspace{-2mm}
\end{figure*}

\noindent \textbf{Illumination estimation framework.} To estimate the illumination of a given image $I$, we first obtain an initial illumination $L'$ by taking the maximum RGB color channels as the illumination value at each pixel~\cite{land1977retinex}, which is expressed as
\begin{equation} \label{equ:init_lum}
L'_p = \max I^c_p,~~\forall c \in \{r, g, b\},
\end{equation}
where $I^c_p$ denotes the color channel $c$ at pixel $p$. The reason why we use the maximal color channel as the initial illumination is that smaller illumination may have the risk of sending color channels of the recovered image $I'$ out of the color gamut, according to $I' = I \times L'^{-1}$. Although the initial illumination map roughly depicts the overall illumination distribution, it typically contains richer details and textures that are not led by illumination discontinuities, making result recovered from it unrealistic; see Figure~\ref{fig:lum_estimation} (b) and (c). Hence, we propose to estimate a refined illumination map $L$ from $L'$ by preserving the prominent structure, while removing the redundant texture details. To this end, we define the following objective function for obtaining the desired illumination map $L$:
\begin{equation} \label{equ:obj_func}
\mathop{\arg \min}\limits_L \sum\limits_p \left( \left (L_p - L'_p\right)^2 + \lambda \left( w_{x,p}\left(\partial_x L\right)_p^2 + w_{y,p}\left(\partial_y L \right)_p^2 \right)\right),
\end{equation}
where $\partial_x$ and $\partial_y$ are spatial derivatives in the horizontal and vertical directions, respectively. $w_{x,p}$ and $w_{y,p}$ are spatially varying smoothness weights. The first term $\left (L_p - L'_p\right)^2$ enforces $L$ to be similar to the initial illumination map $L'$, while the second term aims to remove the redundant texture details in $L'$ by minimizing the partial derivatives. $\lambda$ is a weight for balancing the two terms.  

\begin{figure}
	\centering
	\begin{subfigure}[c]{0.11\textwidth}
		\centering
		\includegraphics[width=0.77in]{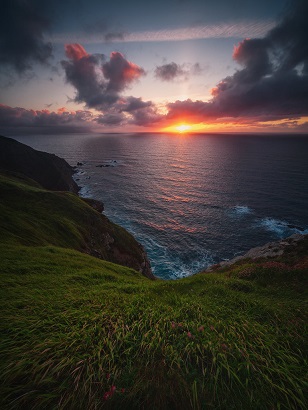}
		\vspace{-1.5em}
		\caption{}
	\end{subfigure}
	\begin{subfigure}[c]{0.11\textwidth}
		\centering
		\includegraphics[width=0.77in]{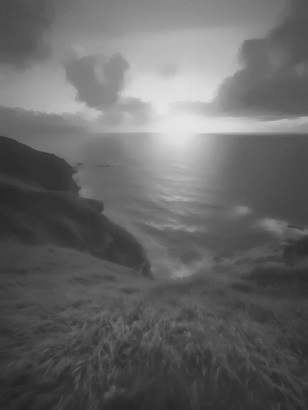}
		\vspace{-1.5em}
		\caption{}
	\end{subfigure}
	\begin{subfigure}[c]{0.11\textwidth}
		\centering
		\includegraphics[width=0.77in]{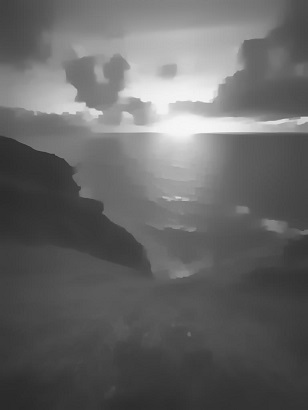}
		\vspace{-1.5em}
		\caption{}
	\end{subfigure}
	\begin{subfigure}[c]{0.11\textwidth}
		\centering
		\includegraphics[width=0.77in]{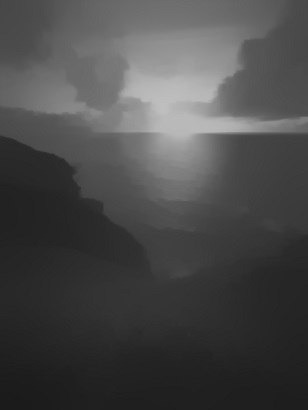}
		\vspace{-1.5em}
		\caption{}
	\end{subfigure} \\ \vspace{1mm}
		\begin{subfigure}[c]{0.11\textwidth}
		\centering
		\includegraphics[width=0.77in]{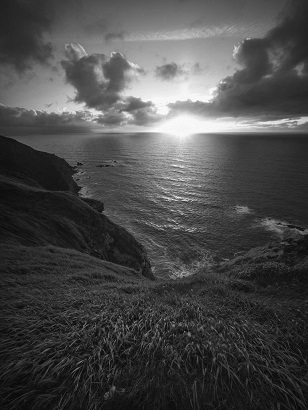}
		\vspace{-1.5em}
		\caption{}
	\end{subfigure}
	\begin{subfigure}[c]{0.11\textwidth}
		\centering
		\includegraphics[width=0.77in]{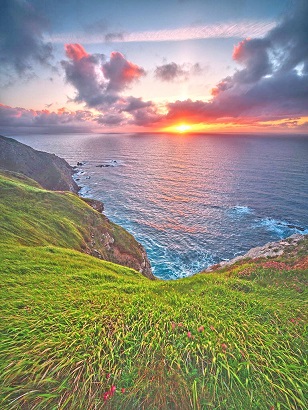}
		\vspace{-1.5em}
		\caption{}
	\end{subfigure}
	\begin{subfigure}[c]{0.11\textwidth}
		\centering
		\includegraphics[width=0.77in]{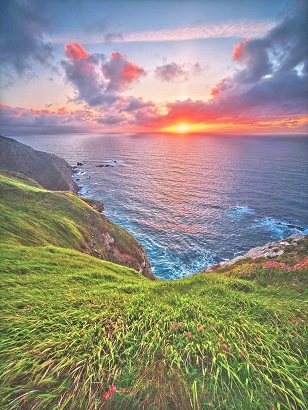}
		\vspace{-1.5em}
		\caption{}
	\end{subfigure}
	\begin{subfigure}[c]{0.11\textwidth}
		\centering
		\includegraphics[width=0.77in]{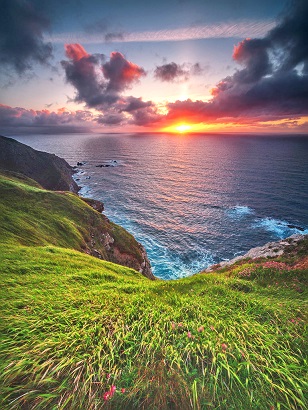}
		\vspace{-1.5em}
		\caption{}
	\end{subfigure}
	\vspace{-2mm}
	\caption{Comparison with edge-preserving smoothing methods on illumination estimation. (a) and (e) are the input image and the initial illumination. (b) and (c) are smoothed illuminations produced by the WLS smoothing~\cite{farbman2008edge} and the RTV method~\cite{xu2012structure}. (f) and (g) are results recovered from the illuminations (b) and (c), respectively. (d) and (h) are our estimated illumination and the corresponding exposure correction result.}
	\label{fig:compare_smoothing}
	\vspace{-2mm}
\end{figure}

Intuitively, the objective function in Eq.~\ref{equ:obj_func} is similar in shape to that of the WLS smoothing~\cite{farbman2008edge}. However, our smoothness weights are defined differently. Specifically, the $x$-direction smoothness weight $w_{x,p}$ is written as
\begin{equation} \label{equ:smooth_weight}
w_{x,p} = \frac{T_{x,p}}{\left| \left(\partial_x L'\right)_p\right| + \epsilon},
\end{equation}
where $T_{x,p}$ is inspired by the relative total variation (RTV) \cite{xu2012structure} and defined as
\begin{equation} \label{equ:rtv_weight}
T_{x,p} = \sum\limits_{q\in\Omega_p}\frac{G_{\sigma}(p,q)}{\left|\sum_{q\in\Omega_p}G_{\sigma}(p,q)(\partial_xL')_q\right|+\epsilon},
\end{equation}
where $\Omega_p$ denotes a $15 \times 15$ squared window centered at pixel $p$. $\epsilon$ in Eqs.~\ref{equ:smooth_weight} and \ref{equ:rtv_weight} are fixed to 1e-3. $G_{\sigma}(p,q)$ computes the spatial affinity based Gaussian weight between pixels $p$ and $q$, and $\sigma = 3$ is the standard deviation. Formally, $G_{\sigma}(p,q)$ is defined as
\begin{equation}
G_{\sigma}(p,q) = \exp\left(-\frac{D(p,q)}{2\sigma^2}\right),
\end{equation} 
where the function $D(p,q)$ computes the spatial Euclidean distance between pixels $p$ and $q$. Since the $y$-direction smoothness weight $U_{y,p}$ is defined similarly, we here do not give its definition. 

\begin{figure*}
	\centering
	\begin{subfigure}[c]{0.19\textwidth}
		\centering
		\includegraphics[width=1.32in]{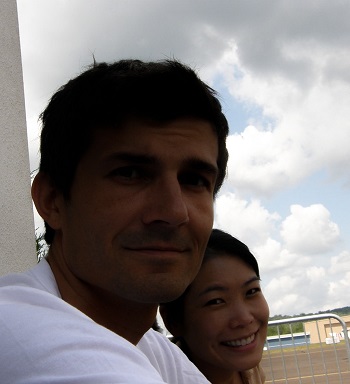}
		\vspace{-1.5em}
		\caption{}
	\end{subfigure}
	\begin{subfigure}[c]{0.19\textwidth}
		\centering
		\includegraphics[width=1.32in]{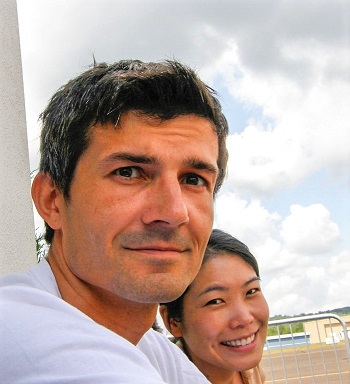}
		\vspace{-1.5em}
		\caption{}
	\end{subfigure}
	\begin{subfigure}[c]{0.19\textwidth}
		\centering
		\includegraphics[width=1.32in]{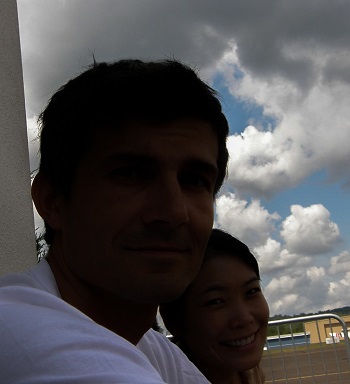}
		\vspace{-1.5em}
		\caption{}
	\end{subfigure}
	\begin{subfigure}[c]{0.19\textwidth}
		\centering
		\includegraphics[width=1.32in]{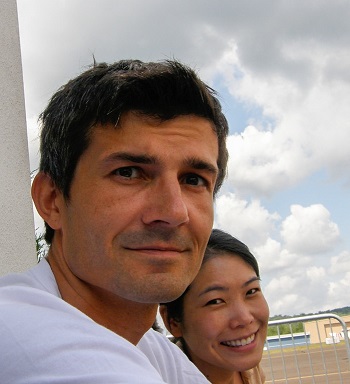}
		\vspace{-1.5em}
		\caption{}
	\end{subfigure}
	\begin{subfigure}[c]{0.19\textwidth}
		\centering
		\includegraphics[width=1.32in]{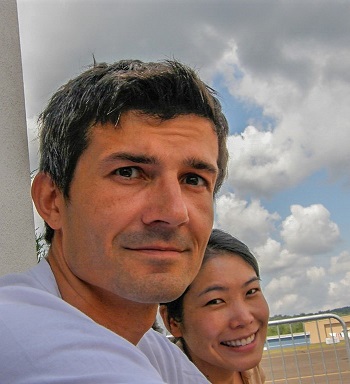}
		\vspace{-1.5em}
		\caption{}
	\end{subfigure}
	\vspace{-2mm}
	\caption{Multi-exposure image fusion. (a) Input image. (b) and (c) are under- and over-exposure corrected images recovered from the forward and reverse illuminations, respectively. (d) and (e) are fused images produced by the original and modified visual quality maps.}
	\label{fig:fusion}
	\vspace{-2mm}
\end{figure*}

The solution to the objective function in Eq.~\ref{equ:obj_func} can be efficiently obtained; see \cite{levin2004colorization,lischinski2006interactive,farbman2008edge} for available solvers. Note that, similar to \cite{fu2016weighted,guo2017lime}, to recover results with better brightness, we alternatively perform a Gamma adjustment to the estimated illumination $L$, \ie, $L = L^{\gamma}$, and recover the exposure correction result by $I' = I * (L^{\gamma})^{-1}$. In our experiments, we empirically set $\gamma$ as 0.6. Figure~\ref{fig:lum_estimation} demonstrates the effectiveness of our illumination estimation in correcting an underexposed image. As can be seen, by optimizing the objective function in Eq.~\ref{equ:obj_func}, we obtain piece-wise smooth illumination with little texture details, from which we recover visually pleasing underexposure correction result. 

Figure~\ref{fig:compare_smoothing} compares our illumination estimation against previous edge-preserving image smoothing methods~\cite{farbman2008edge,xu2012structure}. For fair comparison, we generated their illuminations based on the same initial illumination, using implementations provided by the authors with well-tuned parameters. Moreover, the Gamma adjustment is applied to the illuminations produced by each method when recovering the exposure correction result. As shown, our illumination better removes the redundant texture details in the initial illumination while also preserving the salient illumination structures, and it recovers visually pleasing result with more distinct contrast and more vivid color. Note that, although the forward illumination estimation is performed in Figure~\ref{fig:compare_smoothing}, the above conclusion also holds for the reverse illumination estimation, since the two are built upon the same illumination estimation algorithm. 



\subsection{Multi-exposure image fusion}
As analyzed above, by performing the proposed dual illumination estimation, we can obtain two intermediate exposure correction versions of an input image, with one corrects the underexposed regions and the other one restores the overexposed regions. Intuitively, to generate globally well-exposed image, the key is to seamlessly fuse the locally best exposed parts in the two intermediate exposure correction images. Considering that there may exist normally exposed regions in the input image, we additionally adopt the input image and perform a multi-exposure image fusion on the three images for the final exposure correction result. 


Let $I'_f$ and $I'_r$ denote the intermediate under- and over-exposure corrected images of an input image $I$. We then employ the exposure fusion technique~\cite{mertens2009exposure} to fuse the image sequence $\{I'_f,I'_r,I\}$ into a globally well-exposed image $I'$. Specifically, we first compute a visual quality map for each image in the sequence by:
\begin{equation}
V^k_p = \left(C^k_p\right)^{\beta_C} \times \left(S^k_p\right)^{\beta_S} \times \left(E^k_p\right)^{\beta_E},
\end{equation} 
where $k$ indicates the $k$-th image in the image sequence. $C$, $S$ and $E$ are quantitatively measures for contrast, saturation, and well-exposedness; see \cite{mertens2009exposure} for details. $\beta_C$, $\beta_S$ and $\beta_E$ are parameters for controlling the influence of each measure, which are set to 1 by default. Note, pixels with higher visual quality values are more likely better exposed. The three visual quality maps are then normalized such that they sum up to one at each pixel $p$.

Next, the multi-resolution image fusion technique originated by Burt and Adelson~\cite{burt1983laplacian} is employed to seamlessly blend images in the sequence under the guidance of the pre-computed visual quality maps. Figure~\ref{fig:fusion} shows an example. As shown, the fused image in Figure~\ref{fig:fusion}(d) adaptively keeps the visually best parts in the multi-exposure image sequence (Figure~\ref{fig:fusion}(a)-(c)), and has better visual appeal compared with the input image due to the improved brightness, clear details, distinct contrast and vivid color. However, we notice that there is a clear quality degradation of locally best exposed regions from the image sequence in the fused image, such as the faces and the sky. We found that this is because the influence of these regions are weakened by same regions with lower visual quality in the sequence during the fusion. Hence, instead of normalizing the visual quality maps, we propose to modify the visual quality maps by only keeping the largest value at each pixel along the image sequence, which is expressed as
\begin{equation}
\hat{V}^k_p = 	\left\{ 
{\begin{array}{lc}
	{1,} &  \text{if}~~k = \mathop{\arg \max}_j V^j_p,~\forall j\in [1,3] \\
	{0,} & \text{otherwise}  	  
	\end{array}} \right..
\end{equation}
Using the modified visual quality maps, we obtain an improved result with clearer face and cloud details as well as better contrast and more vivid color, as shown in Figure~\ref{fig:fusion}(e).

\subsection{Implementation and parameter setting}
We implement our algorithm using Matlab on Core i5-7400 CPU 3.0GHz. Similar to \cite{xu2011image}, we alternatively optimize the objective function in Eq.~\ref{equ:obj_func} in the Fourier domain for speedup, which takes 0.3 seconds for estimating the illumination of a 1 mega-pixel image. For the multi-exposure image fusion, we adopt the implementation provided by the authors of \cite{mertens2009exposure}, which takes about 1.5 seconds to produce the result for a 1 mega-pixel image. Note, as analyzed in \cite{mertens2009exposure}, an optimized GPU implementation would greatly accelerate the fusion process and enable real-time performance. Our code will be made publicly available at \url{http://zhangqing-home.net/} .

\begin{figure}
	\centering
	\begin{subfigure}[c]{0.11\textwidth}
		\centering
		\includegraphics[width=0.77in]{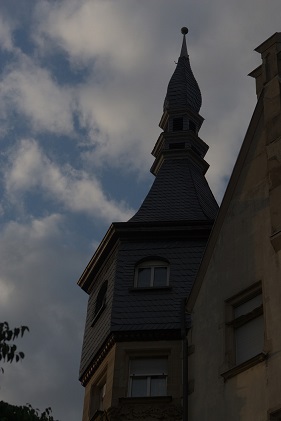}  \\ \vspace{2pt}
		\includegraphics[width=0.77in]{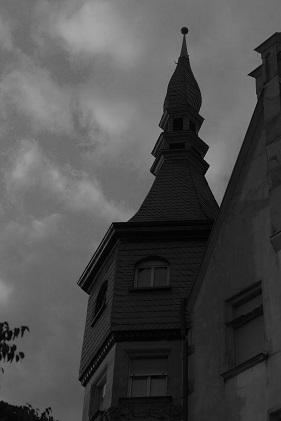} 
		\caption{Input}
	\end{subfigure}
	\begin{subfigure}[c]{0.11\textwidth}
		\centering
		\includegraphics[width=0.77in]{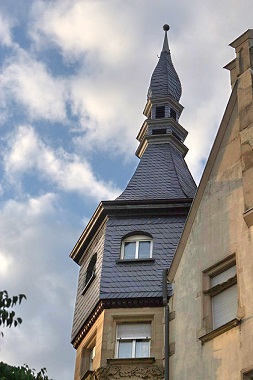} \\ \vspace{2pt}
		\includegraphics[width=0.77in]{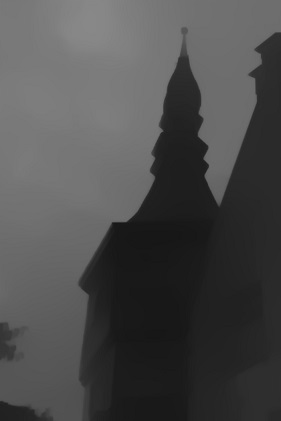}
		\caption{$\lambda=0.1$}
	\end{subfigure}
	\begin{subfigure}[c]{0.11\textwidth}
		\centering
		\includegraphics[width=0.77in]{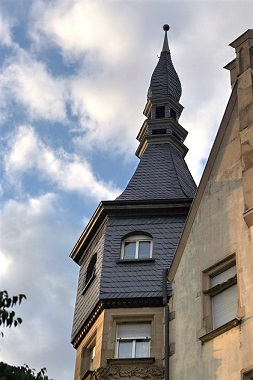} \\ \vspace{2pt}
		\includegraphics[width=0.77in]{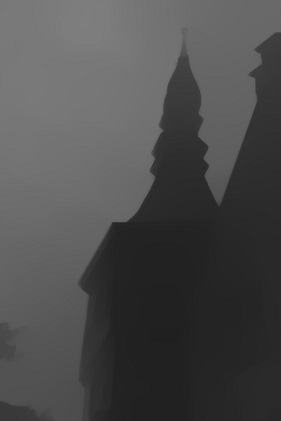}
		\caption{$\lambda = 0.3$}
	\end{subfigure}
	\begin{subfigure}[c]{0.11\textwidth}
		\centering
		\includegraphics[width=0.77in]{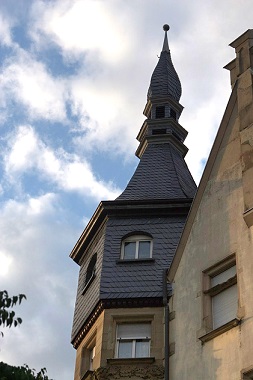} \\ \vspace{2pt}
		\includegraphics[width=0.77in]{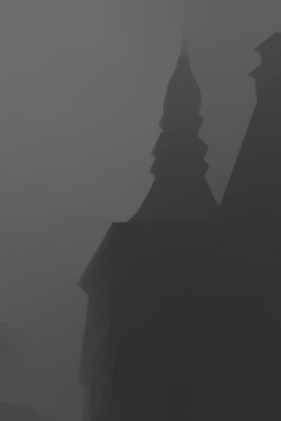}
		\caption{$\lambda=1.2$}
	\end{subfigure}  \vspace{-1mm}
	\caption{Effect of varying $\lambda$. (a) Input image (top) and the initial forward illumination (bottom). (b)-(c) are exposure corrected images and the corresponding forward illuminations. }
	\label{fig:parameters}
	\vspace{-2mm}
\end{figure}

The key parameter in our algorithm is $\lambda$, which controls the smoothness level of the resulting illuminations. In general, larger $\lambda$ yields smoother illumination, which allows recovering exposure corrected image with stronger local contrast. However, an excessively smoothed illumination would in turn decrease the brightness and contrast. To obtain better visual results, we set $\lambda = 0.15$ in all our experiments, which produces good results. Figure~\ref{fig:parameters} shows an example illustrating how $\lambda$ affects the forward illumination and the recovered underexposure corrected image.

\section{Experiments}
To objectively evaluate the effectiveness of our method, we conducted the user study in two aspects. We first conducted a user study to compare our method with existing automatic exposure correction tools, and then conducted another study to compare our method with the state-of-the-art methods. 


\begin{figure*}
	\centering
	\begin{subfigure}[c]{0.19\textwidth}
		\centering
		\includegraphics[width=1.32in]{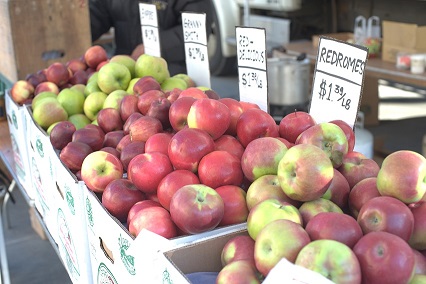} \\ \vspace{2pt}
		\includegraphics[width=1.32in]{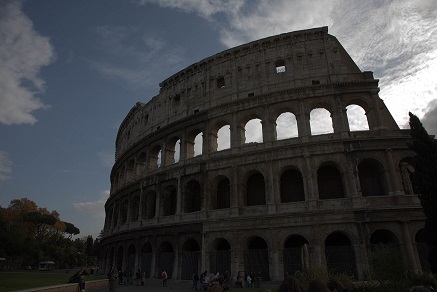} \\ \vspace{2pt}
		\includegraphics[width=1.32in]{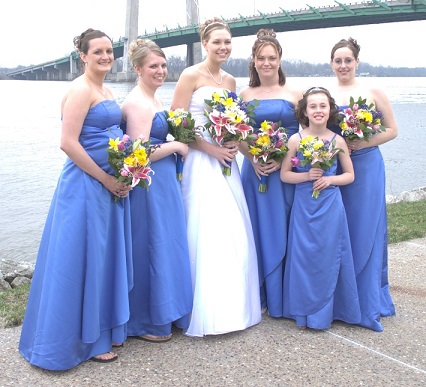} \\ \vspace{2pt}
		\includegraphics[width=1.32in]{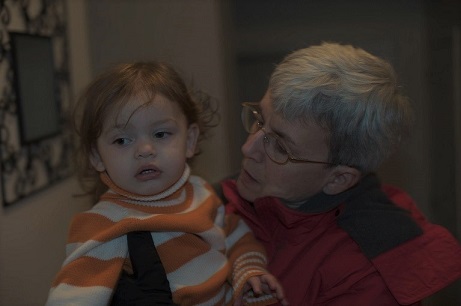} 
		\caption{Input}
	\end{subfigure}
	\begin{subfigure}[c]{0.19\textwidth}
		\centering
		\includegraphics[width=1.32in]{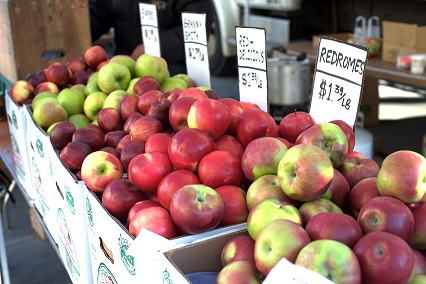} \\ \vspace{2pt}
		\includegraphics[width=1.32in]{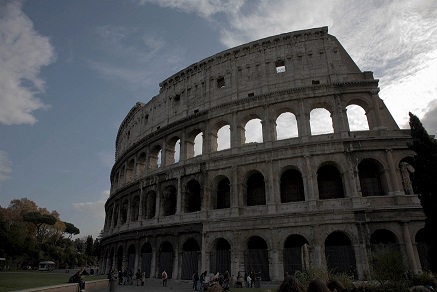} \\ \vspace{2pt}
		\includegraphics[width=1.32in]{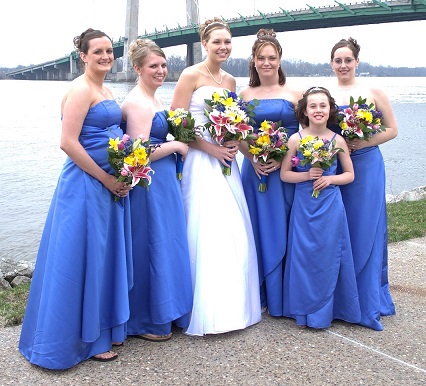}  \\ \vspace{2pt}
		\includegraphics[width=1.32in]{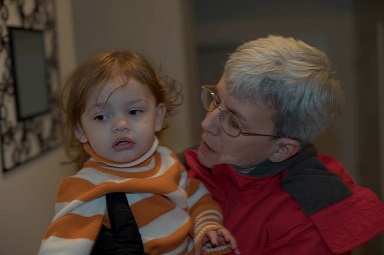} 
		\caption{Auto-Level}
	\end{subfigure}
	\begin{subfigure}[c]{0.19\textwidth}
		\centering
		\includegraphics[width=1.32in]{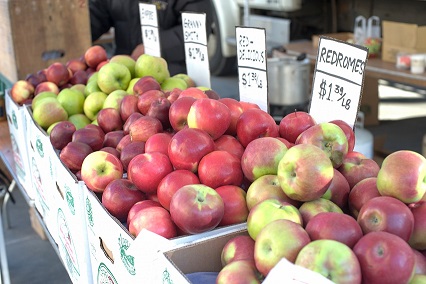} \\ \vspace{2pt}
		\includegraphics[width=1.32in]{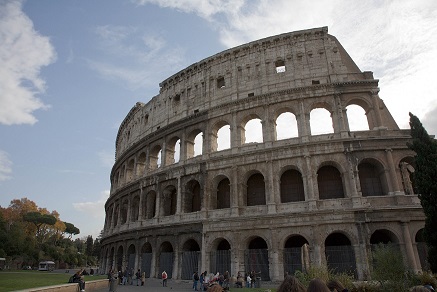}  \\ \vspace{2pt}
		\includegraphics[width=1.32in]{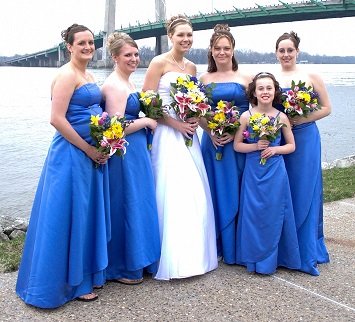}  \\ \vspace{2pt}
		\includegraphics[width=1.32in]{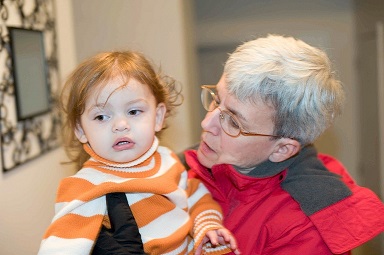}  
		\caption{Auto-Tone}
	\end{subfigure}
	\begin{subfigure}[c]{0.19\textwidth}
		\centering
		\includegraphics[width=1.32in]{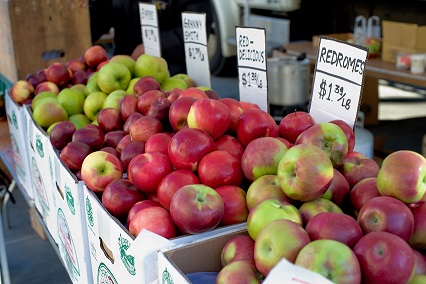} \\ \vspace{2pt}
		\includegraphics[width=1.32in]{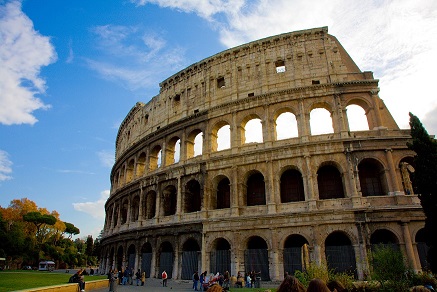} \\ \vspace{2pt}
		\includegraphics[width=1.32in]{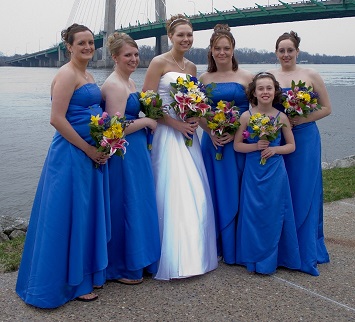} \\ \vspace{2pt}
		\includegraphics[width=1.32in]{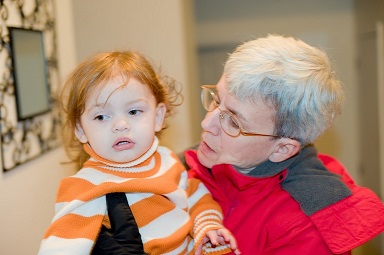} 
		\caption{Expert-retouched}
	\end{subfigure}
	\begin{subfigure}[c]{0.19\textwidth}
		\centering
		\includegraphics[width=1.32in]{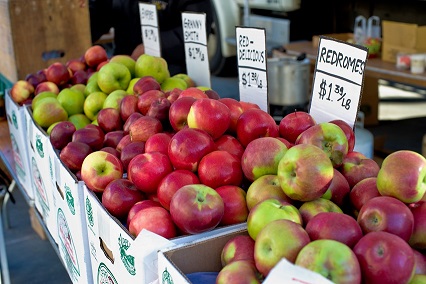} \\ \vspace{2pt}
		\includegraphics[width=1.32in]{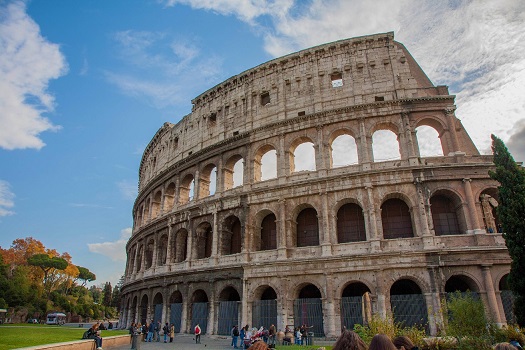}  \\ \vspace{2pt}
		\includegraphics[width=1.32in]{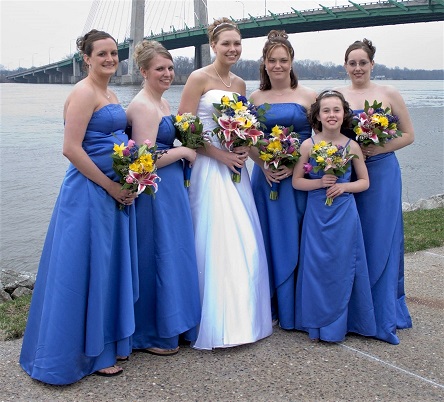}  \\ \vspace{2pt}
		\includegraphics[width=1.32in]{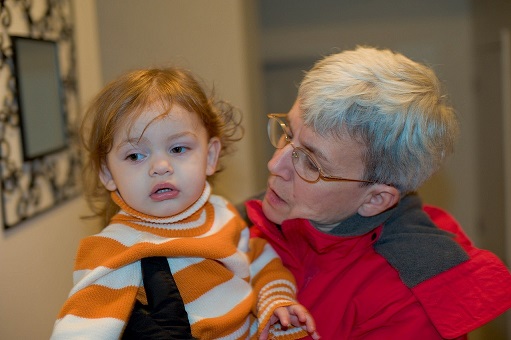} 
		\caption{Our result}
	\end{subfigure}
	\vspace{-2mm}
	\caption{Visual comparison with automatic exposure correction tools and expert retouching on several example images employed in the first user study. The input images and the expert-retouched results are from the MIT-Adobe FiveK dataset~\cite{bychkovsky2011learning}.}
	\label{fig:user_study_1}
	\vspace{-2mm}
\end{figure*}

\subsection{User study \#1: comparison with existing tools}
In this study, we compare our method with two automatic exposure correction tools, including Auto-Level in Photoshop and Auto-Tone in Lightroom, and interactive exposure correction via Lightroom. The two automatic tools and our method are first applied to 100 randomly selected images from the MIT-Adobe FiveK dataset \cite{bychkovsky2011learning}. We then collected the results and recruited 100 participants via Amazon Mechanical Turk to rate their preferences to the results. Similar to \cite{yuan2012automatic,kaufman2012content}, each participant was informed to perform pairwise comparison between our result and one of the four other images: 1) input image, 2) result of Auto-Level, 3) result of Auto-Tone, and 4) the corresponding expert-retouched results from Expert C in the employed dataset, according to the following five common requirements for the desired exposure correction result: (i) suitable brightness, (ii) clear details, (iii) distinct contrast, (iv) vivid color, and (v) well-preserved photorealism. For each pairwise comparison, the participants have three options: ``I prefer the image on the left'', ``I prefer the image on the right'', and ``the two images look the same to me''. Note that, to avoid subjective bias, each image pair was presented anonymous and in a random order throughout the evaluation.

\begin{table}[]
	\begin{tabular}{l|c|c|c}
		\hline
		Methods                    & Other & Same  & Ours  \\ \hline
		Our method vs. Input            & 6\% & 9\% & \textbf{85}\% \\
		Our method vs. Auto-Level       & 7\% & 11\% & \textbf{82}\% \\
		Our method vs. Auto-Tone        & 21\% & 12\% & \textbf{67}\% \\
		Our method vs. Expert-retouched & \textbf{42}\% & 21\% & 37\% \\ \hline
	\end{tabular}
	\caption{User study \#1. ``Ours'': the percentage that our result is preferred. ``Other'': the percentage that the other compared image is preferred. ``Same'': the percentage that the participants have no clear preference to the shown image pair. \textbf{Bold} indicates the best.}
	\label{table:us_1}
	\vspace{-2mm}
\end{table}

Table~\ref{table:us_1} summarizes the statistical results of the pairwise comparison. Comparing the results, we can notice that our method has better performance on the pairwise comparison between the input image, Auto-Level, and Auto-Tone, which convincingly demonstrates that results generated by our method are more preferred by human subjects. In addition, as shown in the last row of the table, the preference of our method is comparable to that of the expert retouching, manifesting that our method is able to produce high-quality exposure correction results and can be a good candidate exposure correction tool for non-expert users. 



Figure~\ref{fig:user_study_1} shows several example images employed in the user study and the exposure correction results generated by the compared tools and our method. As can be seen, the input images are diverse, including (i) a fruit stand image with overexposed apples (1st row), (ii) a landscape image with underexposed building and overexposed sky (2nd row), (iii) an overexposed wedding image with abnormal skin color and unclear details (3rd row), and (iv) a globally underexposed image with little portrait details (4th row). As shown, Auto-Level is less effective in correcting underexposure and fail to restore severely overexposed regions. Compared with Auto-Level, Auto-Tone is relatively more effective, but may also fail to produce satisfactory results due to inherent difficulty of automatically balancing multiple appearance factors. 
In contrast, our method generates visually pleasing results with normal brightness, clear details, distinct contrast and vivid color, which are comparable to the corresponding expert-retouched results. Please see the supplementary material for more image results in the user study.


\subsection{User study \#2: comparison with state-of-the-art methods}
\begin{figure*}
	\centering
	\begin{subfigure}[c]{0.19\textwidth}
		\includegraphics[width=1.32in]{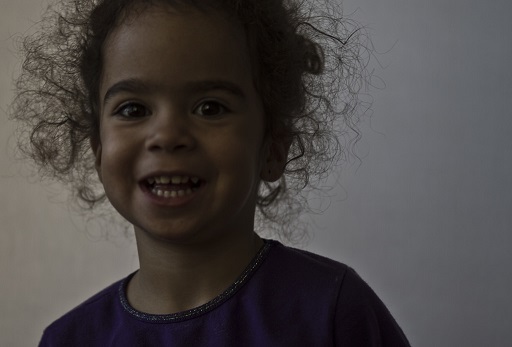} \\ \vspace{-3.5mm}
		\includegraphics[width=1.32in]{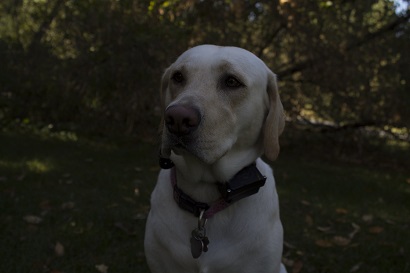} \\ \vspace{-3.5mm}
		\includegraphics[width=1.32in]{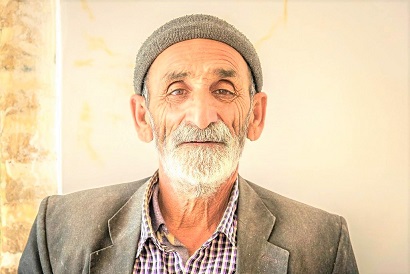} \\ \vspace{-3.5mm}
		\includegraphics[width=1.32in]{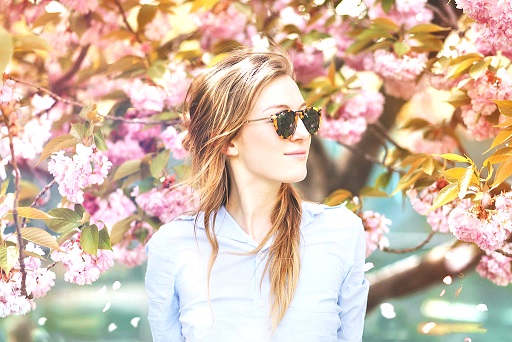}
		\caption{Input}
	\end{subfigure}
	\begin{subfigure}[c]{0.19\textwidth}
		\includegraphics[width=1.32in]{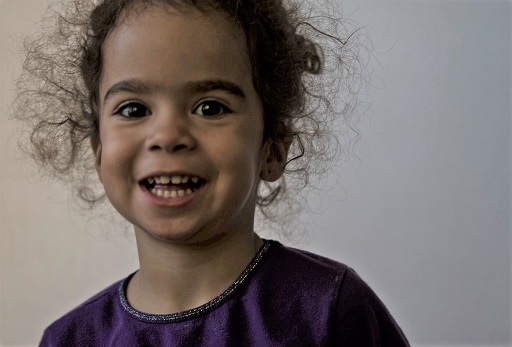} \\  \vspace{-3.5mm}
		\includegraphics[width=1.32in]{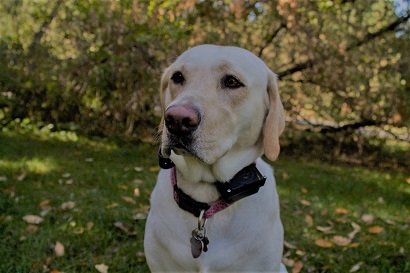} \\ \vspace{-3.5mm}
		\includegraphics[width=1.32in]{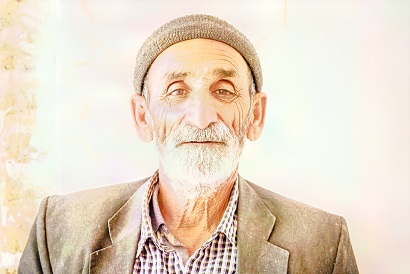} \\ \vspace{-3.5mm}
		\includegraphics[width=1.32in]{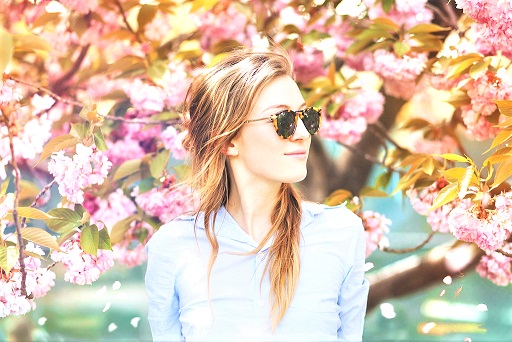}
		\caption{HDRNet~\cite{gharbi2017deep}}
	\end{subfigure}
	\begin{subfigure}[c]{0.19\textwidth}
		\includegraphics[width=1.32in]{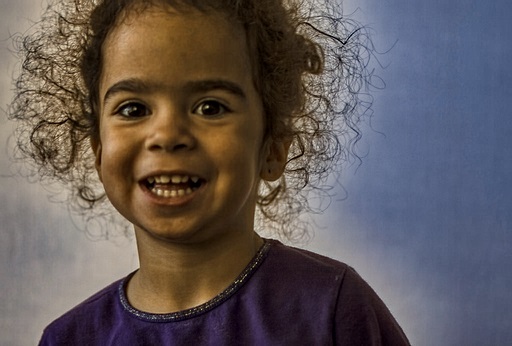} \\  \vspace{-3.5mm}
		\includegraphics[width=1.32in]{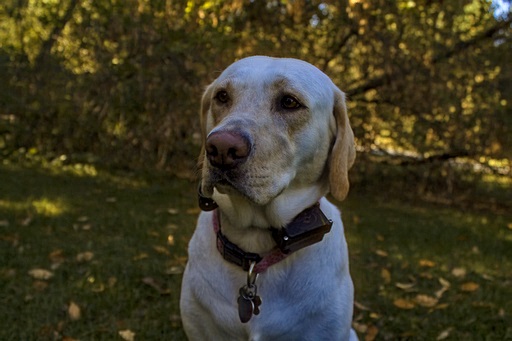} \\ \vspace{-3.5mm}
		\includegraphics[width=1.32in]{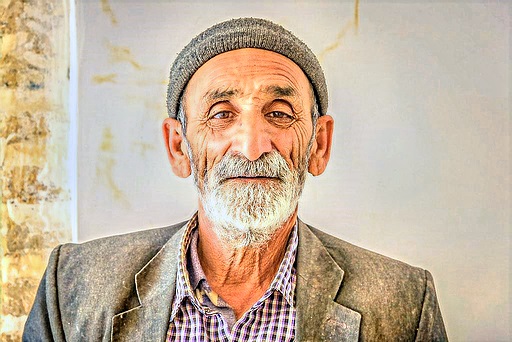} \\ \vspace{-3.5mm}
		\includegraphics[width=1.32in]{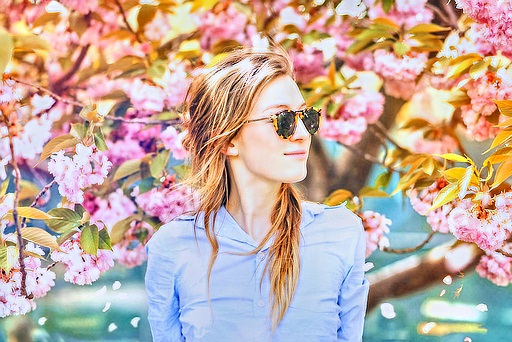}
		\caption{DPE~\cite{chen2018deep}}
	\end{subfigure}
	\begin{subfigure}[c]{0.19\textwidth}
		\includegraphics[width=1.32in]{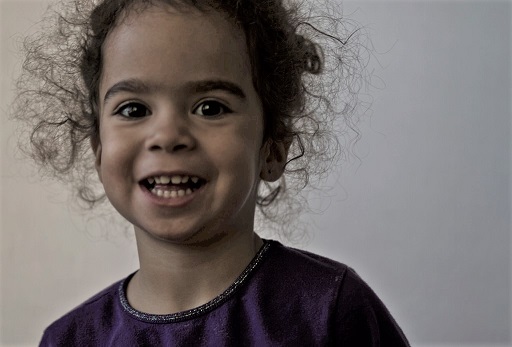} \\  \vspace{-3.5mm}
		\includegraphics[width=1.32in]{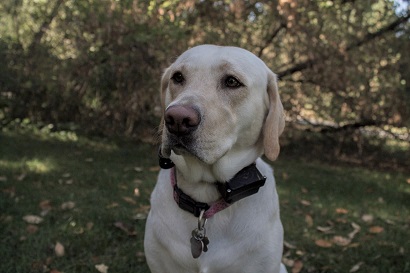} \\ \vspace{-3.5mm}
		\includegraphics[width=1.32in]{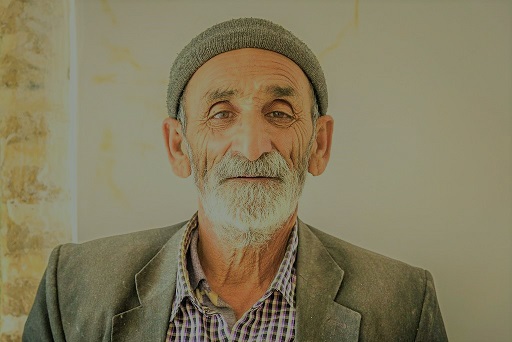} \\ \vspace{-3.5mm}
		\includegraphics[width=1.32in]{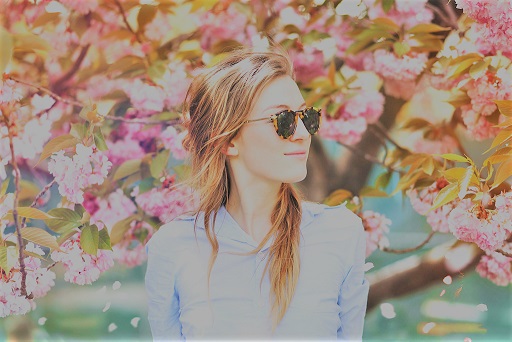}
		\caption{Exposure~\cite{hu2018exposure}}
	\end{subfigure}
	\begin{subfigure}[c]{0.19\textwidth}
		\includegraphics[width=1.32in]{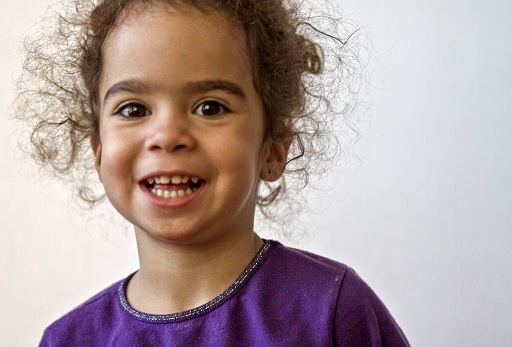} \\ \vspace{-3.5mm}
		\includegraphics[width=1.32in]{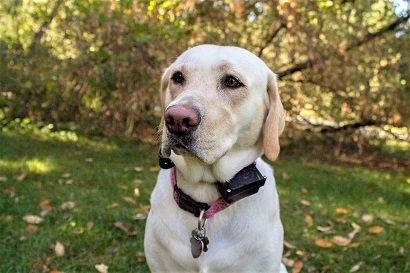} \\ \vspace{-3.5mm}
		\includegraphics[width=1.32in]{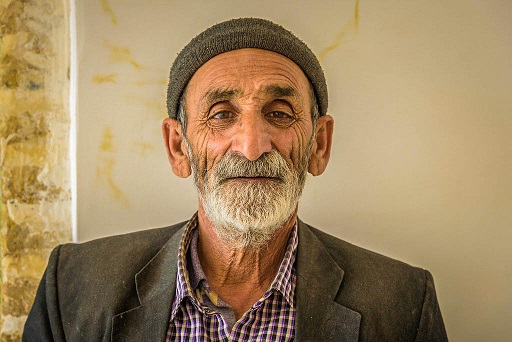} \\ \vspace{-3.5mm}
		\includegraphics[width=1.32in]{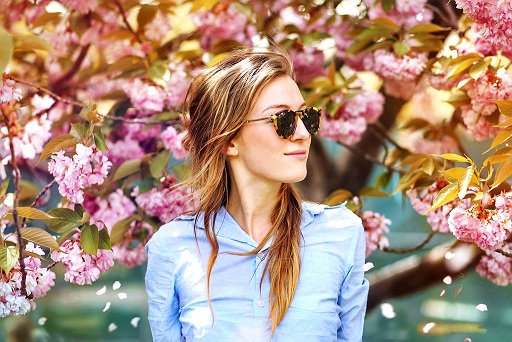}
		\caption{Ours}
	\end{subfigure}
	\vspace{-2mm}
	\caption{Visual comparison with state-of-the-art methods on several example images employed in the second user study.}
	\label{fig:user_study_2}
	\vspace{-2mm}
\end{figure*}

Here we compare our method with three recent learning-based exposure correction methods, including HDRNet~\cite{gharbi2017deep}, DPE~\cite{chen2018deep}, and Exposure~\cite{hu2018exposure}. Since these methods are trained on the MIT-Adobe FiveK dataset, unlike the first user study, we are unable to use the randomly selected images from the MIT-Adobe FiveK dataset for evaluation. Hence, we first
crawled 100 test images that have over 50\% pixels with normalized intensity lower than 0.3 (69 images) or higher than 0.7 (31 images), from Flicker by searching with keywords ``low light'', ``underexposed'', ``overexposed'', ``back light'', and ``portrait''; see Figure~\ref{fig:user_study_2} for examples. Then, the three compared methods and our method were used for exposure correction of the 100 test images. For fair comparison, we produced their results using publicly-available implementation provided by the authors with recommended parameter setting. Next, similar to the first user study, pairwise preference comparison on the Amazon Mechanical Turk with 100 participants were conducted between our result and one of the four images: 1) input image, 2) result of HDRNet~\cite{gharbi2017deep}, 3) result of DPE~\cite{chen2018deep}, and 4) result of Exposure~\cite{hu2018exposure}. 

\begin{table}[]
	\begin{tabular}{l|c|c|c}
	\hline
	Methods                    & Other & Same  & Ours  \\ \hline
	Our method vs. Input            & 6\% & 11\% & \textbf{83}\% \\
	Our method vs. HDRNet~\cite{gharbi2017deep}       & 16\% & 12\% & \textbf{72}\% \\
	Our method vs. DPE~\cite{chen2018deep}        & 13\% & 9\% & \textbf{78}\% \\
	Our method vs. Exposure~\cite{hu2018exposure} & 18\% & 15\% & \textbf{67}\% \\ \hline
\end{tabular}
	\caption{User study \#2. Note, the meanings of ``Ours'', ``Same'', and ``Other'' are same as that in Table~\ref{table:us_1}. \textbf{Bold} indicates the best.}
	\label{table:us_2}
	\vspace{-2mm}
\end{table}


The user study results are shown in Table~\ref{table:us_2}, where we can see that our results were favored by more human subjects, indicating that our method outperforms the compared methods. Figure~\ref{fig:user_study_2} shows some test images and their exposure correction results generated by different methods in the user study. As shown, HDRNet~\cite{gharbi2017deep} erroneously increases the exposure of the overexposed portrait images in the third and fourth rows, leading to visually unpleasing results. DPE~\cite{chen2018deep} is effective in color and contrast enhancement, but may generate unrealistic results. Exposure~\cite{hu2018exposure} produces competitive results for the underexposed images in the first and second rows, while it fails to generate satisfactory results for the last two overexposed images. In comparison, we produce more appealing results. Additional image results in the user study are provided in the supplementary material. 

\subsection{More Analysis}

\begin{figure}
	\centering
	\begin{subfigure}[c]{0.15\textwidth}
		\centering
		\includegraphics[width=1.03in]{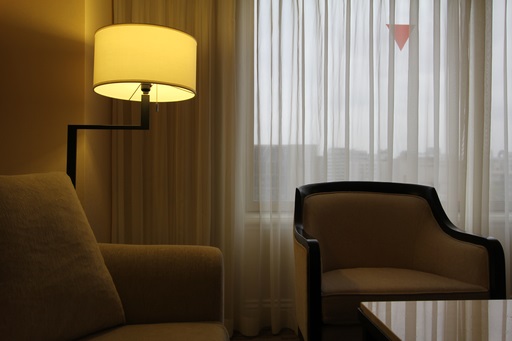}
	\end{subfigure}
	\begin{subfigure}[c]{0.15\textwidth}
		\centering
		\includegraphics[width=1.03in]{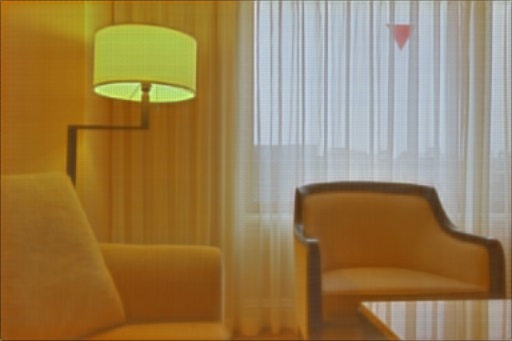}
	\end{subfigure}
	\begin{subfigure}[c]{0.15\textwidth}
		\centering
		\includegraphics[width=1.03in]{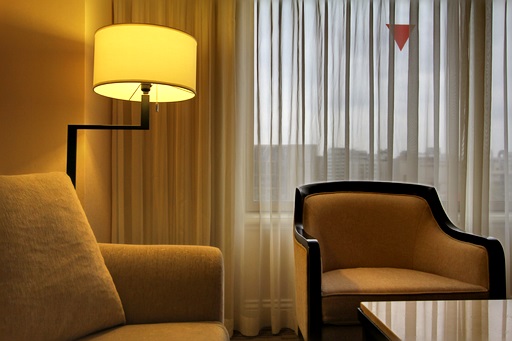}
	\end{subfigure}
	\caption{Comparison with IID. Left to right are the input image, reflectance generated by Li~\etal~\cite{li2018learning}, and our exposure correction result. Source image from the IIW dataset~\cite{bell2014intrinsic}.}
	\vspace{-2mm}
	\label{fig:comp_iid}
\end{figure}

\noindent \textbf{Relationship to intrinsic image decomposition.} Although the assumption in Eq.~\ref{equ:retinex} looks the same as that of the intrinsic image decomposition (IID), our problem is essentially different from IID. In contrast, IID assumes that an image is the pixel-wise product of the material reflectance and the illumination, where the reflectance component corresponds to an unrealistic image independent of illumination, while the same part (\ie, $I'$) in our model is the desired natural-looking exposure correction image. Figure~\ref{fig:comp_iid} shows an example comparing our result with the reflectance component produced by the state-of-the-art IID method. 

\begin{figure}
	\centering
	\begin{subfigure}[c]{0.22\textwidth}
		\centering
		\includegraphics[width=1.52in]{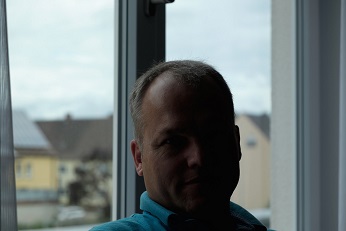} \\ \vspace{2pt}
		\includegraphics[width=1.52in]{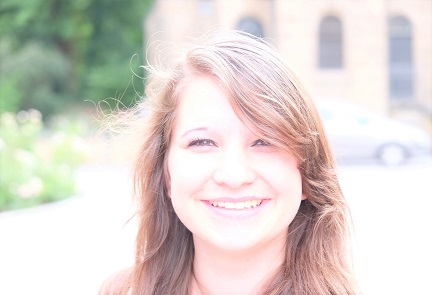}
	\end{subfigure}
	\begin{subfigure}[c]{0.22\textwidth}
		\centering
		\includegraphics[width=1.52in]{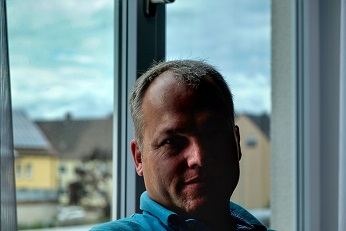} \\ \vspace{2pt}
		\includegraphics[width=1.52in]{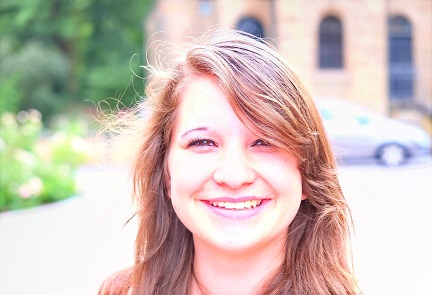}
	\end{subfigure}
	\caption{Failed cases. \textbf{Left:} Input images. \textbf{Right:} Our results. Our method fails to recover missing image contents from the extremely under- and over-exposed facial regions.}
	\label{fig:limitations}
	\vspace{-2mm}
\end{figure}

\noindent \textbf{Limitations.} While our method produces satisfactory results for most of our test images, it still has a few limitations. First, as shown in Figure~\ref{fig:limitations}, our method fails to produce visually compelling results for the two images, since some parts of the facial regions are almost black and white, and without any trace of color and texture in the original images. Note that other state-of-the-art methods also fail to produce satisfactory results for the input images in Figure~\ref{fig:limitations}; see supplementary material for their results. Another limitation is that our method may amplify noise together with the fine scale details when the input image is noisy.

\section{Conclusion and Future Work}
We have presented a novel exposure correction method. Unlike previous methods, we propose to estimate dual illuminations, which allows us to conveniently recover high-quality intermediate under- and over-exposure corrected images. A multi-exposure image fusion technique is then adopted to integrate the locally best exposed parts in the two intermediate exposure correction images and the input image into a globally well-exposed image. Overall, our method is simple yet effective and can run fully automatically in near-interactive rate. We have performed extensive experiments on a number of images and compare our method with popular automatic exposure correction tools and the state-of-the-art methods to demonstrate its effectiveness.

Our future work is threefold. First, we will investigate how to suppress noise during exposure correction. Second, we are interested in adopting semantics information and texture synthesis techniques to recover the missing image content for extremely under- and over-exposed regions as that in Figure~\ref{fig:limitations}. Third, inspired by \cite{zhu2018saliency}, we will try to improve the results by guiding the illumination estimation with saliency.

\noindent \textbf{Acknowledgments~~} The authors thank the reviewers for their
valuable comments. This work was partially supported by
the NSFC (No. U1811461, No. 61802453 and No. 61602183), and the Science and Technology Project of Guangzhou (201707010140).

\bibliographystyle{eg-alpha-doi} 
\bibliography{egbibsample}           

\newcommand{\etalchar}[1]{$^{#1}$}
\begin{thebibliography}{\uppercase{LWWG15}}

\bibitem[ABK{\etalchar{*}}18]{abebe2018towards}
\textsc{Abebe M.~A., Booth A., Kervec J., Pouli T., Larabi M.-C.}:
\newblock Towards an automatic correction of over-exposure in photographs:
  Application to tone-mapping.
\newblock \emph{Computer Vision and Image Understanding 168} (2018), 3--20.

\bibitem[BA83]{burt1983laplacian}
\textsc{Burt P., Adelson E.}:
\newblock The laplacian pyramid as a compact image code.
\newblock \emph{IEEE Transactions on Communications 31}, 4 (1983), 532--540.

\bibitem[BBS14]{bell2014intrinsic}
\textsc{Bell S., Bala K., Snavely N.}:
\newblock Intrinsic images in the wild.
\newblock \emph{ACM Transactions on Graphics 33}, 4 (2014), 159.

\bibitem[BM05]{bennett2005video}
\textsc{Bennett E.~P., McMillan L.}:
\newblock Video enhancement using per-pixel virtual exposures.
\newblock \emph{ACM Transactions on Graphics 24}, 3 (2005), 845--852.

\bibitem[BPCD11]{bychkovsky2011learning}
\textsc{Bychkovsky V., Paris S., Chan E., Durand F.}:
\newblock Learning photographic global tonal adjustment with a database of
  input/output image pairs.
\newblock In \emph{CVPR} (2011), pp.~97--104.

\bibitem[CGZ18]{cai2018learning}
\textsc{Cai J., Gu S., Zhang L.}:
\newblock Learning a deep single image contrast enhancer from multi-exposure
  images.
\newblock \emph{IEEE Transactions on Image Processing 27}, 4 (2018),
  2049--2062.

\bibitem[CWKC18]{chen2018deep}
\textsc{Chen Y.-S., Wang Y.-C., Kao M.-H., Chuang Y.-Y.}:
\newblock Deep photo enhancer: Unpaired learning for image enhancement from
  photographs with gans.
\newblock In \emph{CVPR} (2018), pp.~6306--6314.

\bibitem[DGV09]{dodgson2009contrast}
\textsc{Dodgson N.~A., Grundland M., Vohra R.}:
\newblock Contrast brushes: Interactive image enhancement by direct
  manipulation.
\newblock \emph{Computational Aesthetics} (2009), 107--114.

\bibitem[DJS{\etalchar{*}}09]{dale2009image}
\textsc{Dale K., Johnson M.~K., Sunkavalli K., Matusik W., Pfister H.}:
\newblock Image restoration using online photo collections.
\newblock In \emph{ICCV} (2009), pp.~2217--2224.

\bibitem[DWP{\etalchar{*}}11]{dong2011fast}
\textsc{Dong X., Wang G., Pang Y., Li W., Wen J., Meng W., Lu Y.}:
\newblock Fast efficient algorithm for enhancement of low lighting video.
\newblock In \emph{ICME} (2011), IEEE, pp.~1--6.

\bibitem[EKD{\etalchar{*}}17]{eilertsen2017hdr}
\textsc{Eilertsen G., Kronander J., Denes G., Mantiuk R.~K., Unger J.}:
\newblock Hdr image reconstruction from a single exposure using deep cnns.
\newblock \emph{ACM Transactions on Graphics 36}, 6 (2017), 178.

\bibitem[EKM17]{endo2017deep}
\textsc{Endo Y., Kanamori Y., Mitani J.}:
\newblock Deep reverse tone mapping.
\newblock \emph{ACM Transactions on Graphics 36}, 6 (2017), 177--1.

\bibitem[FFLS08]{farbman2008edge}
\textsc{Farbman Z., Fattal R., Lischinski D., Szeliski R.}:
\newblock Edge-preserving decompositions for multi-scale tone and detail
  manipulation.
\newblock \emph{ACM Transactions on Graphics 27}, 3 (2008), 67.

\bibitem[FZH{\etalchar{*}}16]{fu2016weighted}
\textsc{Fu X., Zeng D., Huang Y., Zhang X.-P., Ding X.}:
\newblock A weighted variational model for simultaneous reflectance and
  illumination estimation.
\newblock In \emph{CVPR} (2016), pp.~2782--2790.

\bibitem[GCB{\etalchar{*}}17]{gharbi2017deep}
\textsc{Gharbi M., Chen J., Barron J.~T., Hasinoff S.~W., Durand F.}:
\newblock Deep bilateral learning for real-time image enhancement.
\newblock \emph{ACM Transactions on Graphics 36}, 4 (2017), 118.

\bibitem[GCZS10]{guo2010correcting}
\textsc{Guo D., Cheng Y., Zhuo S., Sim T.}:
\newblock Correcting over-exposure in photographs.
\newblock In \emph{CVPR} (2010), pp.~515--521.

\bibitem[GLL17]{guo2017lime}
\textsc{Guo X., Li Y., Ling H.}:
\newblock Lime: Low-light image enhancement via illumination map estimation.
\newblock \emph{IEEE Transactions on Image Processing 26}, 2 (2017), 982--993.

\bibitem[HHX{\etalchar{*}}18]{hu2018exposure}
\textsc{Hu Y., He H., Xu C., Wang B., Lin S.}:
\newblock Exposure: A white-box photo post-processing framework.
\newblock \emph{ACM Transactions on Graphics 37}, 2 (2018), 26.

\bibitem[HJS13]{hou2013recovering}
\textsc{Hou L., Ji H., Shen Z.}:
\newblock Recovering over-/underexposed regions in photographs.
\newblock \emph{SIAM Journal on Imaging Sciences 6}, 4 (2013), 2213--2235.

\bibitem[JMAK10]{joshi2010personal}
\textsc{Joshi N., Matusik W., Adelson E.~H., Kriegman D.~J.}:
\newblock Personal photo enhancement using example images.
\newblock \emph{ACM Transactions on Graphics 29}, 2 (2010), 12--1.

\bibitem[Kim97]{kim1997contrast}
\textsc{Kim Y.-T.}:
\newblock Contrast enhancement using brightness preserving bi-histogram
  equalization.
\newblock \emph{IEEE Transactions on Consumer Electronics 43}, 1 (1997), 1--8.

\bibitem[KLW12]{kaufman2012content}
\textsc{Kaufman L., Lischinski D., Werman M.}:
\newblock Content-aware automatic photo enhancement.
\newblock \emph{Computer Graphics Forum 31}, 8 (2012), 2528--2540.

\bibitem[Lan77]{land1977retinex}
\textsc{Land E.~H.}:
\newblock The retinex theory of color vision.
\newblock \emph{Scientific American 237}, 6 (1977), 108--129.

\bibitem[LFUS06]{lischinski2006interactive}
\textsc{Lischinski D., Farbman Z., Uyttendaele M., Szeliski R.}:
\newblock Interactive local adjustment of tonal values.
\newblock \emph{ACM Transactions on Graphics 25}, 3 (2006), 646--653.

\bibitem[LLW04]{levin2004colorization}
\textsc{Levin A., Lischinski D., Weiss Y.}:
\newblock Colorization using optimization.
\newblock \emph{ACM transactions on graphics 23}, 3 (2004), 689--694.

\bibitem[LS18]{li2018learning}
\textsc{Li Z., Snavely N.}:
\newblock Learning intrinsic image decomposition from watching the world.
\newblock In \emph{CVPR} (2018), pp.~9039--9048.

\bibitem[LWWG15]{li2015low}
\textsc{Li L., Wang R., Wang W., Gao W.}:
\newblock A low-light image enhancement method for both denoising and contrast
  enlarging.
\newblock In \emph{ICIP} (2015), IEEE, pp.~3730--3734.

\bibitem[LYKK14]{lee2014correction}
\textsc{Lee D.-H., Yoon Y.-J., Kang S.-j., Ko S.-J.}:
\newblock Correction of the overexposed region in digital color image.
\newblock \emph{IEEE Transactions on Consumer Electronics 60}, 2 (2014),
  173--178.

\bibitem[MG16]{masia2016content}
\textsc{Masia B., Gutierrez D.}:
\newblock Content-aware reverse tone mapping.
\newblock In \emph{ICAITA} (2016), Atlantis Press.

\bibitem[MKVR09]{mertens2009exposure}
\textsc{Mertens T., Kautz J., Van~Reeth F.}:
\newblock Exposure fusion: A simple and practical alternative to high dynamic
  range photography.
\newblock \emph{Computer Graphics Forum 28}, 1 (2009), 161--171.

\bibitem[PLYSK18]{park2018distort}
\textsc{Park J., Lee J.-Y., Yoo D., So~Kweon I.}:
\newblock Distort-and-recover: Color enhancement using deep reinforcement
  learning.
\newblock In \emph{CVPR} (2018), pp.~5928--5936.

\bibitem[RSSF02]{reinhard2002photographic}
\textsc{Reinhard E., Stark M., Shirley P., Ferwerda J.}:
\newblock Photographic tone reproduction for digital images.
\newblock \emph{ACM transactions on graphics 21}, 3 (2002), 267--276.

\bibitem[Sta00]{stark2000adaptive}
\textsc{Stark J.~A.}:
\newblock Adaptive image contrast enhancement using generalizations of
  histogram equalization.
\newblock \emph{IEEE Transactions on Image Processing 9}, 5 (2000), 889--896.

\bibitem[WZF{\etalchar{*}}19]{wang2019underexposed}
\textsc{Wang R., Zhang Q., Fu C.-W., Shen X., Zheng W.-S., Jia J.}:
\newblock Underexposed photo enhancement using deep illumination estimation.
\newblock In \emph{CVPR} (2019), pp.~6849--6857.

\bibitem[WZHL13]{wang2013naturalness}
\textsc{Wang S., Zheng J., Hu H.-M., Li B.}:
\newblock Naturalness preserved enhancement algorithm for non-uniform
  illumination images.
\newblock \emph{IEEE Transactions on Image Processing 22}, 9 (2013),
  3538--3548.

\bibitem[XLXJ11]{xu2011image}
\textsc{Xu L., Lu C., Xu Y., Jia J.}:
\newblock Image smoothing via l0 gradient minimization.
\newblock \emph{ACM Transactions on Graphics 30}, 6 (2011), 174.

\bibitem[XYXJ12]{xu2012structure}
\textsc{Xu L., Yan Q., Xia Y., Jia J.}:
\newblock Structure extraction from texture via relative total variation.
\newblock \emph{ACM Transactions on Graphics 31}, 6 (2012), 139.

\bibitem[YS12]{yuan2012automatic}
\textsc{Yuan L., Sun J.}:
\newblock Automatic exposure correction of consumer photographs.
\newblock In \emph{ECCV} (2012), pp.~771--785.

\bibitem[YZW{\etalchar{*}}16]{yan2016automatic}
\textsc{Yan Z., Zhang H., Wang B., Paris S., Yu Y.}:
\newblock Automatic photo adjustment using deep neural networks.
\newblock \emph{ACM Transactions on Graphics 35}, 2 (2016), 11.

\bibitem[ZHF{\etalchar{*}}18]{zhu2018saliency}
\textsc{Zhu L., Hu X., Fu C.-W., Qin J., Heng P.-A.}:
\newblock Saliency-aware texture smoothing.
\newblock \emph{IEEE Transactions on Visualization and Computer Graphics}
  (2018).

\bibitem[ZNZX16]{zhang2016underexposed}
\textsc{Zhang Q., Nie Y., Zhang L., Xiao C.}:
\newblock Underexposed video enhancement via perception-driven progressive
  fusion.
\newblock \emph{IEEE Transactions on Visualization and Computer Graphics 22}, 6
  (2016), 1773--1785.

\bibitem[Zui94]{zuiderveld1994contrast}
\textsc{Zuiderveld K.}:
\newblock Contrast limited adaptive histogram equalization.
\newblock In \emph{Graphics gems IV} (1994), pp.~474--485.

\bibitem[ZYX{\etalchar{*}}18]{zhang2018high}
\textsc{Zhang Q., Yuan G., Xiao C., Zhu L., Zheng W.-S.}:
\newblock High-quality exposure correction of underexposed photos.
\newblock In \emph{ACM Multimedia} (2018), pp.~582--590.

\end{thebibliography}




\end{document}